\documentclass[letterpaper, 10 pt, conference]{ieeeconf}  

\IEEEoverridecommandlockouts                              

\let\labelindent\relax

\usepackage{amsmath} %
\usepackage{amssymb}  %
\usepackage{hyperref}
\hypersetup{
    colorlinks=true,
    linkcolor=black,
    citecolor=black,
    filecolor=cyan,
    urlcolor=blue
}
\usepackage{multirow}
\usepackage{booktabs}
\usepackage{siunitx}
\usepackage{url}
\usepackage{multirow}
\usepackage{dirtytalk}
\usepackage{cite}
\usepackage{enumitem}
\usepackage[font={small}]{caption}
\usepackage{subcaption}

\usepackage[usenames,dvipsnames]{xcolor}
\definecolor{Gray1}{gray}{0.0}
\definecolor{Gray2}{gray}{0.1}
\definecolor{Gray3}{gray}{0.25}
\definecolor{Gray4}{gray}{0.4}

\usepackage{misc/onimage}
\tikzset{
    image label/.style={
        every node/.style={
            fill=black,
            text=white,
            font=\fontfamily{phv}\selectfont\small\bfseries,
            anchor=south west,
            xshift=0.1cm,
            yshift=0.1cm,
            at={(0,0)}
        }
    }
}

\newcommand{\etal}{\textit{et al}. }

\newcommand{\secref}[1]{Sec.~\ref{#1}}

\newcommand{\figref}[1]{Fig.~\ref{#1}}
\newcommand{\tabref}[1]{Table~\ref{#1}}

\usepackage[normalem]{ulem}

\usepackage[colorinlistoftodos]{todonotes}

\usepackage{accents}

\newcommand{\ve}{\mathbf e}
\newcommand{\vf}{\mathbf f}
\newcommand{\vg}{\mathbf g}
\newcommand{\vh}{\mathbf h}

\newcommand{\vp}{\mathbf p}

\newcommand{\vu}{\mathbf u}

\newcommand{\vx}{\mathbf x}

\newcommand{\vQ}{\mathbf Q}
\newcommand{\vR}{\mathbf R}

\title{
Articulated Object Interaction in Unknown Scenes\\ with Whole-Body Mobile Manipulation 
}

\author{
Mayank Mittal$^{\dagger\ast}$, 
David Hoeller$^{\dagger \ast}$, 
Farbod Farshidian$^{\dagger}$,
Marco Hutter$^{\dagger}$, 
Animesh Garg$^{\ddagger \ast}$ 
\thanks{This research was supported by the Swiss National Science Foundation through the National Centre of Competence in Digital Fabrication (NCCR dfab), CIFAR AI Chair, and NSERC Discovery Award.}
\thanks{$^\dagger$ M. Mittal, D. Hoeller, F. Farshidian and M. Hutter are with ETH Z\"{u}rich, Switzerland. {Email: \tt\footnotesize{mittalma}@ethz.ch}}
\thanks{$^\ddagger$ A. Garg is with Vector Institute and University of Toronto, Canada.}
\thanks{$^\ast$ M. Mittal, D. Hoeller and A. Garg are also with NVIDIA.}
}

\begin{document}

\bstctlcite{IEEEexample:BSTcontrol}

\maketitle

\begin{abstract}

A kitchen assistant needs to operate human-scale objects, such as cabinets and ovens, in unmapped environments with dynamic obstacles. 
Autonomous interactions in such environments require integrating dexterous manipulation and fluid mobility. 
While mobile manipulators in different form factors provide an extended workspace, their real-world adoption has been limited. 
Executing a high-level task for general objects requires a perceptual understanding of the object as well as adaptive whole-body control among dynamic obstacles.
In this paper, we propose a two-stage architecture for autonomous interaction with large articulated objects in unknown environments. 
The first stage, \emph{object-centric} planner, only focuses on the object to provide an action-conditional sequence of states for manipulation using RGB-D data.
The second stage, \emph{agent-centric} planner,  formulates the whole-body motion control as an optimal control problem that ensures safe tracking of the generated plan, even in scenes with moving obstacles.
We show that the proposed pipeline can handle complex static and dynamic kitchen settings for both wheel-based and legged mobile manipulators.
Compared to other agent-centric planners, our proposed planner achieves a higher success rate and a lower execution time.
We also perform hardware tests on a legged mobile manipulator to interact with various articulated objects in a kitchen. 
For additional material, please check: \href{https://www.pair.toronto.edu/articulated-mm/}{www.pair.toronto.edu/articulated-mm/}.

\end{abstract}

\section{Introduction}

Unlike in factories, where the environments can be designed to fit a robot's needs and be modeled \textit{a priori}, deploying autonomous service robots into human environments, such as households, is highly challenging~\cite{kemp2007challenges}. There exist dramatic variations in these environments, such as in the room's architecture or objects' placements.
Even within an object category, there are differences in an object's shape, size, and appearance. Thus, to interact successfully with an object instance in an unstructured environment, it is essential to reason over its properties and to have real-time capabilities for handling dynamic changes.

\begin{figure}[t]
    \vspace{10pt}
    \centering
    \includegraphics[trim=15 0 15 0, clip, width=0.9\linewidth]{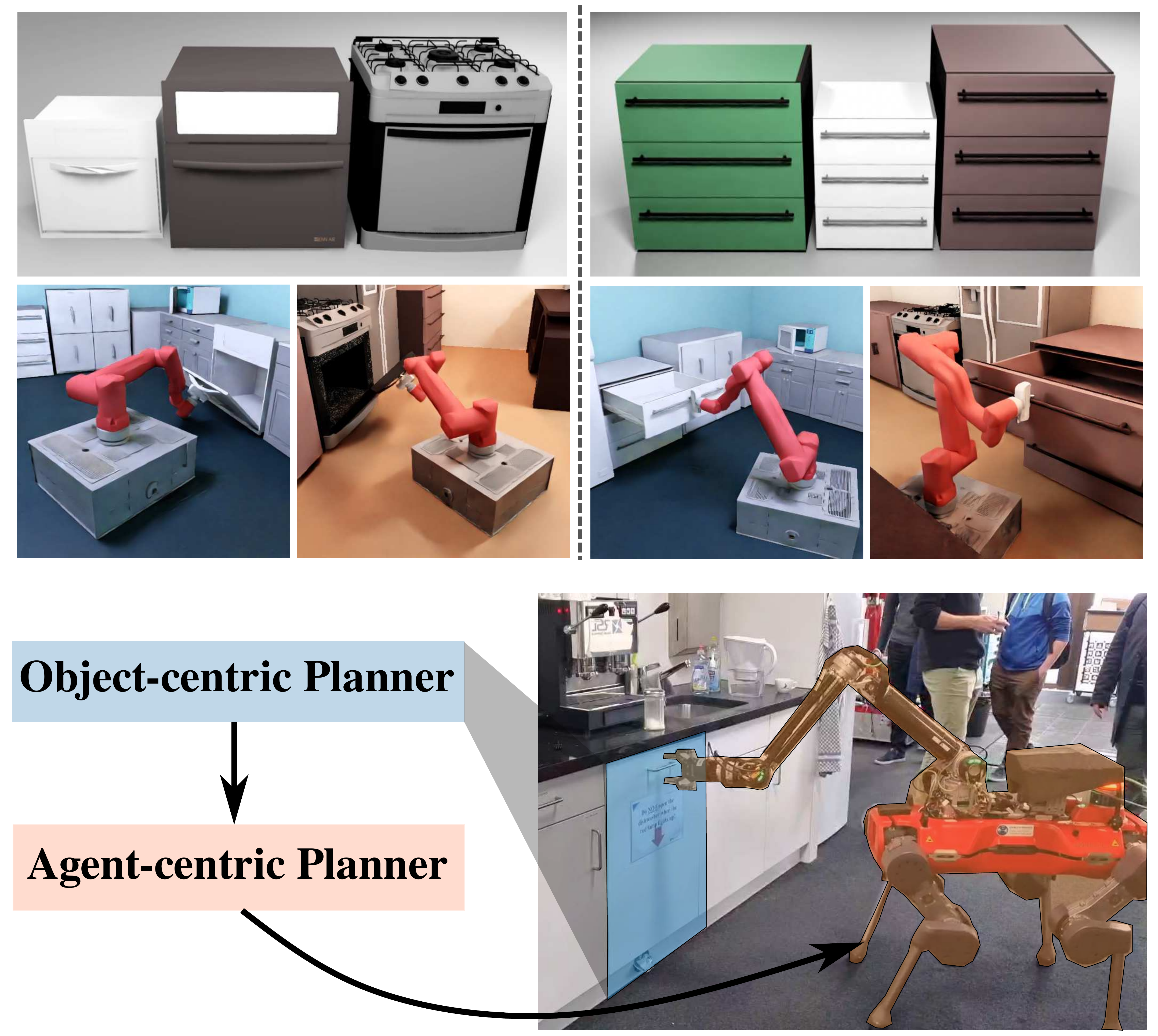}
    \caption{
    We present a framework for articulated object interaction that can handle unknown spaces, variations within an object category, and dynamic scenes. We consider both wheel-based and legged mobile manipulation systems.
    }
    \label{fig:teaser}
\end{figure}

We propose an intuitive framework for the manipulation of articulated objects in novel scenes. We decompose the decision-making into a two-level planning hierarchy-- \textit{object-centric} and \textit{agent-centric}. In this formulation, the object-centric planner provides proposals for interacting with the object without considering scene information or the agent embodiment
\cite{Li2020ancsh,Welschehold2017humandemo,wu2021vat,mo2021where2act}. The agent-centric planner, on the other hand, ensures safe execution of the proposed plans by the robot and accounts for the robot's dynamics and its environment~\cite{berenson2011tsr,Burget2013wbmparticulated,Pankert2020perceptivempc,sleiman2021unified}. 
For a given task, while a large candidate set of object-centric plans may exist, their feasibility depends on the robot's physical capabilities (e.g., robot's joint limits), the environment (e.g., cluttering in the workspace), and the range of motions induced by the agent-centric planner.

In this paper, we benchmark choices for agent-centric planners to maximize the feasibility of successful articulated objects interaction in unknown environments. Particularly, we look at three different planners for mobile manipulators that use map information for environment collision avoidance: i) sampling-based planning for the base with inverse kinematics (IK) for the arm, ii) whole-body control using IK, and iii) whole-body control via model predictive control (MPC). 
We observe that the MPC-based solution performs significantly better than IK-based solutions for articulated object manipulation in terms of success rate and execution time. 
Since, in this paper, our primary focus is agent-centric planning, we use existing perception methods to compute object-centric plans for manipulation, 
thereby providing a complete mobile manipulation system that can be deployed in the real-world.

Our key contributions are as follows:
\begin{enumerate}[
    noitemsep,
    ]
    \item We adopt and extend our previous collision-free whole-body MPC for a legged system~\cite{gaertner2021collision} to mobile manipulation in unknown static and dynamic scenes. %
    \item Building upon perception methods for object-centric planning and our MPC-based agent-centric planner, we show that the resulting system can handle complex kitchen settings and ensure safe execution for wheel-based and legged mobile manipulators.
    \item We benchmark different agent-centric planners for mobile manipulation of large articulated objects in hyper-realistic simulation. Compared to IK-based solutions, we observe that MPC improves the success rate by $134\%$ and reduces the time taken by $26.5\%$.
    \item We present hardware experiments on a legged mobile manipulator and demonstrate the proposed system's capability to manipulate various articulated objects in a real kitchen.
\end{enumerate}

\section{Related Work}
\label{sec:rel_work}

Robotic manipulation of articulated objects has been researched for more than a decade. While an exhaustive survey is beyond the scope of this article, we mention a few recent efforts with respect to above-listed contributions.

\subsection{Agent-centric Planning for Mobile Manipulators}

Berenson~\etal~\cite{berenson2011tsr} propose a sampling-based planner, based on rapidly exploring random trees (RRTs), for manipulators under end-effector pose constraints. However, their method inherits the shortcomings of RRTs, such as incorporating non-holonomic constraints and dynamics.
To make planning tractable for non-holonomic systems, typically, the base navigation and arm manipulation planning are done separately~\cite{Meeussen2010dooropening,arduengo2019robust}. An essential aspect of such a method is positioning the base to maximize the arm's reachability. While inverse reachability maps~\cite{Vahrenkamp2013irm} help resolve this issue for the initial base placement, it does not consider the base movement during the object manipulation itself.

To coordinate the base and arm during object interaction, recently, policies trained through reinforcement learning either define sub-goals for the base and arm planners and decide which planner to execute~\cite{xia2020relmogen, li2020hrl4in}, or generate kinematically feasible base movement commands~\cite{honerkamp2021learning}. However, as highlighted in~\cite{honerkamp2021learning}, this decoupling can sometimes lead to kinematic failures during fast motions and grasp failures. %

Previous works on motion planning for mobile manipulation of articulated objects~\cite{Burget2013wbmparticulated, chitta2010dooropen} incorporate the object's kinematic model as a task space constraint and apply sampling-based planners with IK to find feasible solutions. However, these methods are often computationally expensive, require offline planning, and may suffer from the shortcomings of IK when approaching singularities.

On the other hand, optimal control provides a framework to incorporate various constraints, perform real-time planning, and not suffer from singularities~\cite{Farshidian2017brealtimeplan}. Recent works on using this formulation for mobile manipulation have shown that a preview horizon in planning yields more robust and better solutions for operational space tracking~\cite{Pankert2020perceptivempc, sleiman2021unified,chiu2022collision}. %

\subsection{Object-centric Planning for Articulated Objects}

Manipulating articulated objects requires reasoning over their properties, often through partially-observed information such as images. 
A common approach is estimating the object's kinematic structure from passive observations~\cite{abbatematteo2019learning,Li2020ancsh,Jain2020ScrewNet,zeng2020visual} or through interactive perception~\cite{MartinMartin2016visualpercepobj,hausman2015active}. These works have been widely used in robotics to infer the articulated object model and incorporate it into planning.

Imitation learning leverages demonstrations from an expert or a teacher~\cite{Welschehold2017humandemo,wong2022error,xiong2021learning}. However, these works require collecting diverse demonstrations, which is often time-consuming and expensive. Instead, more recent works learn \emph{actionable} visual priors for articulated objects from direct interactions~\cite{xu2022umpnet,wu2021vat,mo2021where2act}. These approaches help exploit the geometric and semantic features of the object (e.g.: edges, holes, handles) beyond expert-defined strategies.

Learning-based methods for articulated object interaction are often studied in isolation with a single object. These approaches complement our framework since they provide object-centric plans, while we want to maximize the feasibility of the generated plans on a real robot.

\subsection{Holisitic Systems for Articulated Object Interaction}
A real system for interaction requires a combination of navigation, grasping, articulation model estimation, and motion planning.
Meeussen~\etal~\cite{Meeussen2010dooropening} integrate a navigation system that performs vision and laser-based handle detection and opens doors using a compliant arm. They plan the base motion independently based on heuristics to prevent collisions with the door. 
Arduengo~\etal\cite{arduengo2019robust} propose a system for door opening with a CNN-based door handle detector and a probabilistic kinematic model estimator by observing the door's motion. For motion planning, they use tasks-space regions~\cite{berenson2011tsr} which limits their planner to holonomic systems. 
R\"{u}hr~\etal\cite{ruhr2012openingdoors} present a framework for operating unknown doors and drawers in kitchen spaces, which learns and stores articulation models in a semantic map for future retrieval. They keep the arm fixed during manipulation and use a position-based impedance controller.

\section{Method Overview}
\label{sec:problem}
\label{sec:method}

\begin{figure}
    \centering
    \includegraphics[width=0.975\linewidth]{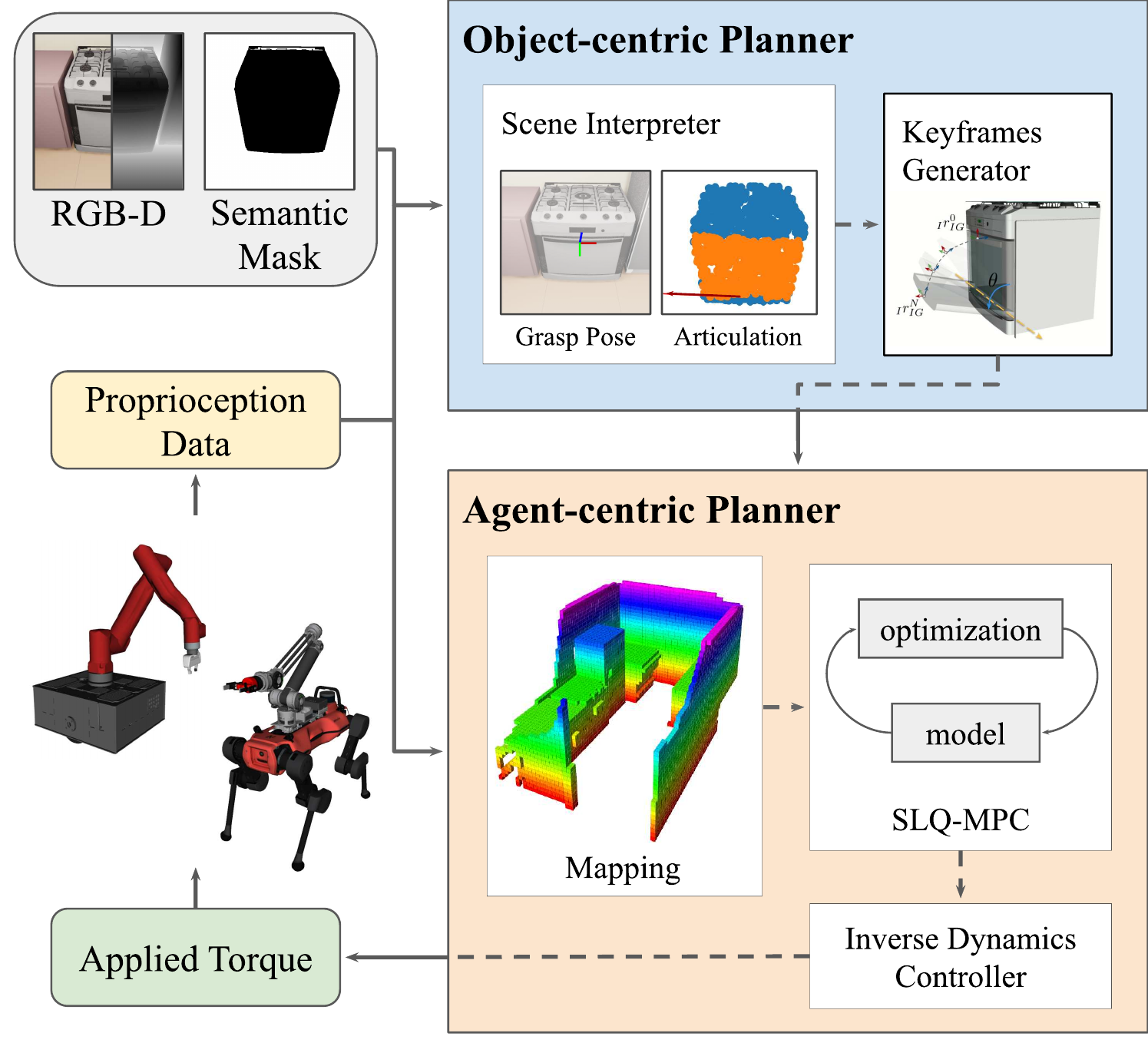}
    \caption{The two-level hierarchy in the proposed framework. The object-centric planner comprises of a scene interpreter and keyframe generator. It uses perceptual information to generate task space plans. The agent-centric planner follows the computed plan while satisfying constraints and performing online collision avoidance.}
    \label{fig:method}
\end{figure}

Object manipulation can be decomposed into planning over two different abstractions: \emph{object-centric} and \emph{agent-centric} (shown in~\figref{fig:method}). Given a certain task, an object-centric planner only generates a plan for an object without considering scene-level information or robot's dynamics. This facilitates generalization of the object-centric planner by avoiding learning of separate representations based on the scene, robot's configuration and its embodiment.

On the other hand, an agent-centric planner, void of object-relevant information, only describes the robot's motions. It may contain information about the robot's model and store environment information through a map representation. Every robot has a set of constraints that defines its capability for safe operation, including its dynamics, joint limits, self- and environment collision.  
The agent-centric planner must operate in this safe region while pushing the operation boundaries to the full range of motions possible.

\begin{figure}
    \centering
    \includegraphics[width=0.8\linewidth]{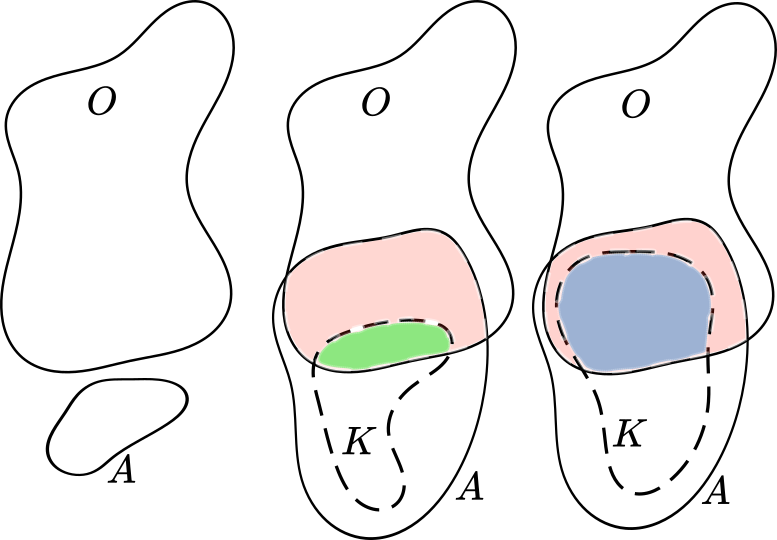}
    \caption{We desire an agent-centric planner ($K$) that maximizes the feasibility of object-centric plans ($O$) for a given robot $A$. \textbf{Left:} When the robot is physically incapable of manipulating the object. \textbf{Middle:} When the robot is capable, but the agent-centric planner limits viable motions. \textbf{Right:} Ideal scenario where the agent-centric planner maximizes the feasible set as much as possible.}
    \label{fig:feasibility_sets}
\end{figure}

Once object-centric plans are computed, they are passed to the agent-centric planner for execution. However, while a large candidate set of object manipulation plans (${O}$) may exist for a given task, their feasibility depends on the robot's physical capabilities (${A}$) and the agent-centric planner (${K}$).
For instance, due to its limited workspace, a fixed-base manipulator may not work for interacting with large articulated objects. For such human environments, a seamless integration of mobility and manipulation is essential.
However, the complexity imposed by higher degrees of freedom (DoF) and the various constraints make the motion generation of mobile manipulators a challenging problem. Thus, the agent-centric planner also influences the candidate object-centric plans that are executable by the robot. This is also illustrated via~\figref{fig:feasibility_sets}.

\noindent\textit{Remarks}. The above separation between object and agent-centric planners does not strictly hold in practice. The robot's end-effector tool heavily determines the possible modes of interaction. For instance, a legged robot's foot is only suitable for non-prehensile actions such as pushing, while a parallel jaw gripper enables both prehensile and non-prehensile actions. Thus, we believe an object-centric planner should be conditioned on the available end-effector tools. This will make the proposed set of candidate plans more sensible for robot execution, and the definition of the agent-centric planner for deciding between interaction options and the motion generation will hold as before. Since this paper only considers a parallel-jaw gripper for grasping, we exclude the conditioning of the object-centric planner from our analysis and only consider object-centric plans on the object's handle.

In this work, we assume that the robot has access to a semantic map of the scene. %
This enables the robot to localize the target object, thereby not requiring to explore an unknown scene to find the object.
We describe particular choices for the object-centric and agent-centric planners in~\secref{sec:task_scheduler} and~\secref{sec:wbmpc} respectively.

\section{Object-centric Planner}
\label{sec:task_scheduler}
Manipulating articulated objects requires a detailed understanding of their parameters (such as joint type, position, and axis orientation) and state. Instead of considering all possible modes of interaction~\cite{wu2021vat,mo2021where2act}, we focus on the most common way, i.e., by grasping the articulated object's handle.

\paragraph{Scene Interpreter} 
We utilize Articulation-aware Normalized Coordinate Space Hierarchy (ANCSH) representation~\cite{Li2020ancsh} to extract articulated object's properties from a pointcloud of the object. Given the articulated object category, the method transforms the input into canonical representations of the object and its parts and estimates part-wise segmentation, joint state, and joint parameters of the object. To infer handles on the object, we use instance segmentation masks.

\paragraph{Keyframe Generation} 
After obtaining the articulation parameters, a grasp pose on the object's handle, and the desired object's state, we compute a uniformly-discretized reference kinematic plan for manipulating the object. The kinematic plan is obtainable for a rotational joint by rotating the grasp frame around the estimated joint axis. For a prismatic joint, the plan is obtainable through translation along the joint axis.

\section{Agent-centric Planner}
\label{sec:wbmpc}

An agent-centric planner for mobile manipulation should ensure: i) tight coordination of the base and arm
, ii) operation within the system's safety limits, iii) reactivity to deal with moving obstacles, and iv) online scene representation to deal with unknown spaces. We design such a planner through a holistic optimization formulation and use a signed distance field (SDF) representation of the world which is generated online using onboard sensors.

In particular, we employ an MPC scheme where at each time step, $t_s$, the MPC solver receives a state measurement, $\vx_s$ and a local map of the environment. Based on these observations, MPC calculates the optimal control action sequence that drives the predicted system output to the desired reference. The robot's low-level control only executes the first piece of this optimized control sequence until the next MPC update arrives. 
We consider the following non-linear optimal control problem where the goal is to find the continuous control input in the time horizon $t_h = t_f - t_s$:
\begin{subequations}
\begin{align}
& \underset{\vu(\cdot)}{\text{minimize}}
&& \phi(\vx(t_s)) + \int_{t_s}^{t_f} l(\vx(t),\vu(t),t) \,\text{d}t, &&
\label{eq:general_cost} 
\\
& \text{subject to} && \dot{\vx} = \vf(\vx,\vu,t),  \quad \vx(t_s) = \vx_s, &&
\label{eq:general_dynamics} 
\\
& && \vg(\vx,\vu,t) = \mathbf{0}, \quad \vh(\vx,\vu,t) \geq \mathbf{0},  &&
\label{eq:general_constraints}
\end{align}
\label{eq:op}%
\end{subequations}
where $\vx(t)$ is the state and $\vu(t)$ is the input at time $t$. $l$ is a time-varying intermediate cost, and $\phi$ is the final cost. The optimization is subjected to the initial condition and system dynamics $\vf$~\eqref{eq:general_dynamics}, and path equality $\vg$ and inequality constraints $\vh$~\eqref{eq:general_constraints}.
In our implementation, we use the Sequential Linear Quadratic (SLQ) method~\cite{Farshidian2017efficoptplan} and employ a real-time iteration scheme for the MPC loop~\cite{Farshidian2017brealtimeplan}. 

We show the effectiveness of this formulation on two different mobile manipulators: wheel-based and legged. We consider a kinematic model for a wheel-based mobile manipulator with a differential drive base with supporting passive castor wheels. The base can turn in place but has a non-holonomic constraint and cannot drive sideways. For a more complex legged mobile manipulator, the system model comprises the centroidal dynamics model and contact constraints that depend on a pre-defined gait schedule. A full description of the models for the wheel-based and legged mobile manipulators are in~\cite{Pankert2020perceptivempc} and~\cite{sleiman2021unified} respectively.

In the following, we describe components of the MPC formulation that are common to any mobile manipulator: end-effector tracking cost function (\secref{sec:cost}), joint constraints (\secref{sec:joint_constraint}), and collision avoidance constraints (\secref{sec:env_constraint}).

\subsection{Task Cost Model}
\label{sec:cost}

We employ the cost function for tasks defined in the robot's end-effector space. In general, the MPC formulation allows planning over pose, twist, and wrench of each end-effector. However, in this work, we only focus on the end-effector pose planning capability of our approach. Consequently, the task optimization variables per end-effector are in $SE(3)$ and the intermediate cost in Eq.~\eqref{eq:op} is written as:
\begin{align}
    &l 
    = 
    \frac{1}{2} \Vert \vu \Vert^2_{\vR_{u}} +
    \sum_{i} 
    \Vert \vp^{i}_{IE} - \widetilde{\vp}^{i}_{IE} \Vert^2_{\vQ^{i}_{p}} 
    + \Vert {\ve_o}^{i}_{IE} \Vert^2_{\vQ^{i}_{o}},
    \label{eq:cost_ee} 
\end{align}
where for the sake of brevity, the dependencies on state, input and time are dropped. The matrices $\vQ^{i}_{p}$ and $\vQ^{i}_{o}$  weights the position and orientation errors respectively at each end-effector and are positive semi-definite. The matrix $\vR_{u}$ penalizes high joint and base velocities and is a positive definite matrix.
In Eq.~\eqref{eq:cost_ee}, $\vp^{i}_{IE}, \widetilde{\vp}^{i}_{IE} \in \mathbb{R}^3$ are the measured and desired positions of the $i^{\textrm{th}}$ end-effector w.r.t. the inertial frame, $I$. $\ve^i_o \in \mathbb{R}^3 $ defines the deviation between the desired and measured orientation through the box minus operator. 

\subsection{Joint Constraints}
\label{sec:joint_constraint}
Linear constraints, which restricts the joint positions of the arm and the velocity commands to the robot, are considered as soft constraints, and penalized through a relaxed barrier function (RBF)~\cite{feller2016relaxed}.%

\subsection{Collision Avoidance Constraints}
\label{sec:env_constraint}

While tracking the end-effector, the generated motion plans for the robot should also prevent collisions with itself and with the environment. In mathematical form, the collision avoidance constraint can be written as:
\begin{equation}
\label{eq:IneqDistConstraint}
    h_j(\vx, t) = d_j(\vx,t) - \epsilon_j \geq 0,
\end{equation}
where $d_j(\vx,t)$ computes the distance between the $j$-th collision pair and $\epsilon_j$ is the minimum distance threshold for each collision pair. The solver handles the inequality path constraints as soft constraints through penalty functions.

\subsubsection{\bf Self-collision avoidance} 
We represent the robot with primitive collision bodies and use GJK algorithm for distance queries through \emph{Flexible Collision Library} (FCL)~\cite{pan2012fcl}. For each pre-specified collision pair $j$ on the robot, we acquire the shortest distance $d_j$ between nearest points on the collision bodies. A negative sign is assigned to the distance if the bodies are colliding. We set the distance threshold $\epsilon_j=0.05$~\si{m}. The constraint is penalized using RBF.

For additional details on distance querying and gradient computation, we refer the readers to~\cite{chiu2022collision}. 

\subsubsection{\bf Environment collision avoidance} 
We incorporate Euclidean SDF (ESDF) for defining an environment collision cost. Since we mainly perform local planning with the MPC, we require a local ESDF around the robot. To this end, we adapt the volumetric-mapping framework~\emph{FIESTA}~\cite{Han2019fiesta}, such that it maintains a globally-consistent occupancy map but updates only a robot-centric ESDF map. 
The approach fuses pointclouds and robot odometry into an occupancy map. From this map, the ESDF is derived using an efficient distance transform algorithm~\cite{Felzenszwalb2012dtsf}.
We further modify FIESTA to support robot-centric \emph{VoxelMap} (similar to GridMap~\cite{Fankhauser2016GridMapLibrary}). This easily allows operations such as tri-linear interpolation to reduce the signed distance error induced by discretizing the map and gradients computation using finite differences. 

\begin{figure}
    \centering
    \begin{subfigure}{0.46\linewidth}
        \centering
        \includegraphics[width=0.95\textwidth, trim=0 0 0 20]{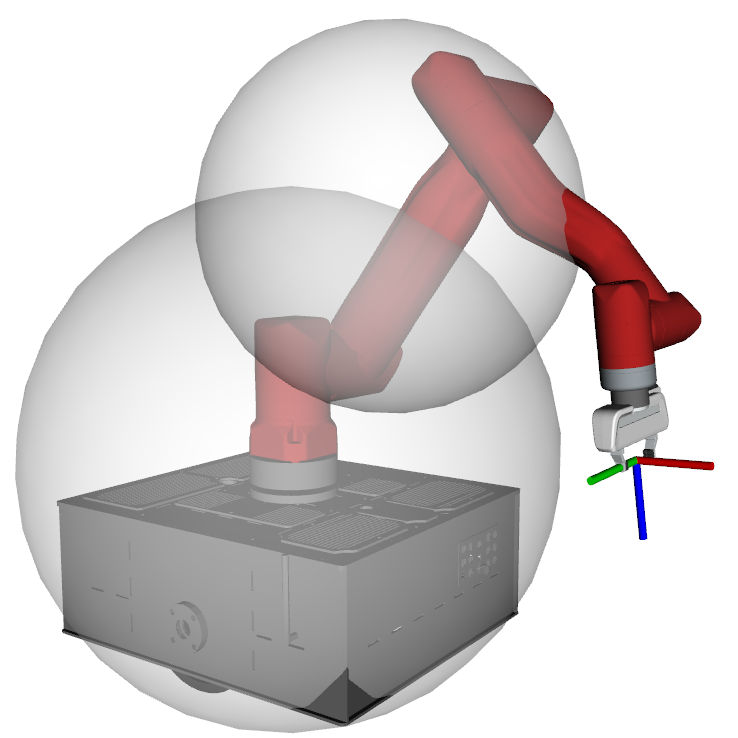}
        \label{fig:mabi}
    \end{subfigure}
    \hfill
    \begin{subfigure}{0.46\linewidth}
        \centering
        \includegraphics[width=0.85\textwidth]{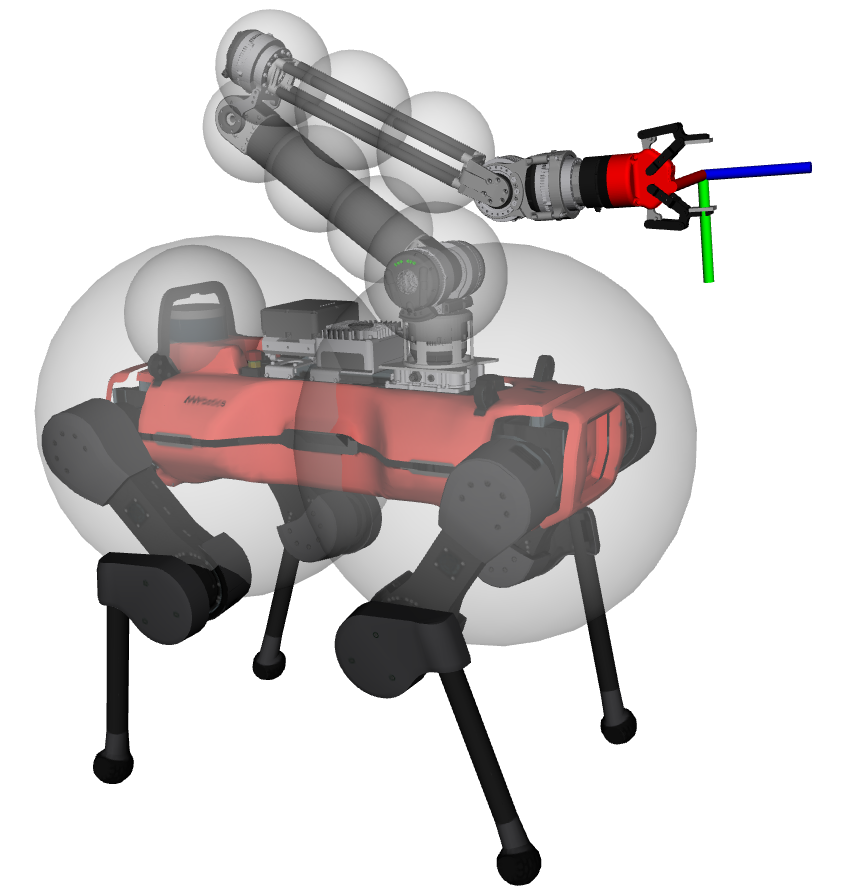}
        \label{fig:alma}
    \end{subfigure}
    \caption{Mobile manipulation platforms considered in this work: (a) \textit{Mabi-Mobile} with a wheel-base, (b) \textit{ALMA} with a legged-base. For collision checking, we approximate the robot with collision spheres. }
    \label{fig:collision_spheres}
    \vspace{-2pt}
\end{figure}

To perform environment collision checking, we define collision spheres to approximate the robot, shown in~\figref{fig:collision_spheres}. The position of each collision sphere $j$ of radius $r_j$ is computed using forward kinematics of the robot $\textrm{FK}_j^I(\vx, t)$. From each center, we query the cached distance and gradient information from the ESDF. Each sphere contributes one collision constraint~\eqref{eq:IneqDistConstraint} with the distance function $d_j$ and threshold $\epsilon_j$ defined as: 
\begin{equation}
d_j(\vx, t) = \textrm{SDF} \left( \textrm{FK}_j^I(\vx, t) \right), \quad \epsilon_j =  r_j,
\end{equation}
where $\textrm{SDF}(\cdot)$ denotes the interpolated distance from ESDF.
We use a Squared-hinge function for the environment collision avoidance constraint $z_j := h_j(\vx, t) \geq 0$ defined as ${H}(z_j) = \frac{\mu}{2} \left( \min\{0, z_j - \delta\} \right)^2$ with $\mu$ and $\delta$ as its hyperparameters.

\section{Experiment Setup}
\label{sec:exp}

In this section , we first introduce the simulation and hardware setups.
We then describe the baselines and performance metrics for evaluating the agent-centric planners.

\subsection{Simulation setup}
\label{sec:sim_platform}

\begin{figure}
    \centering
    \includegraphics[width=0.342\linewidth]{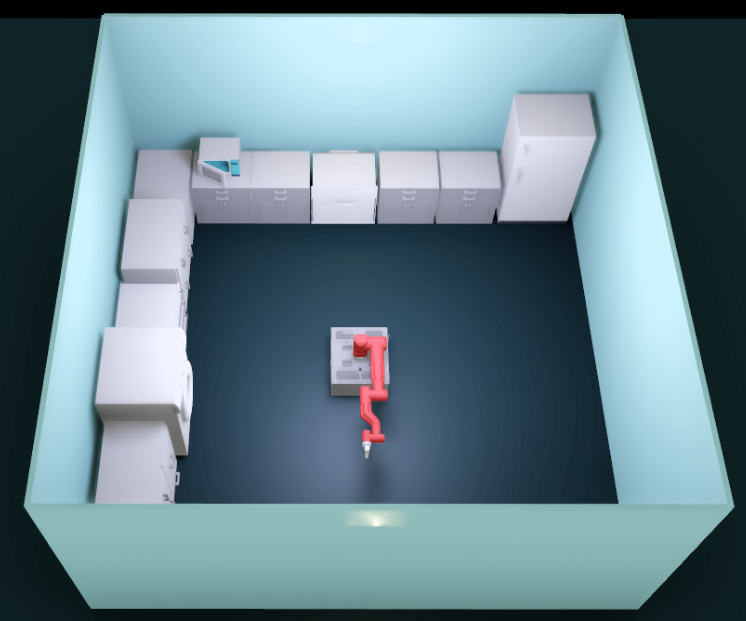}
    \includegraphics[width=0.38\linewidth]{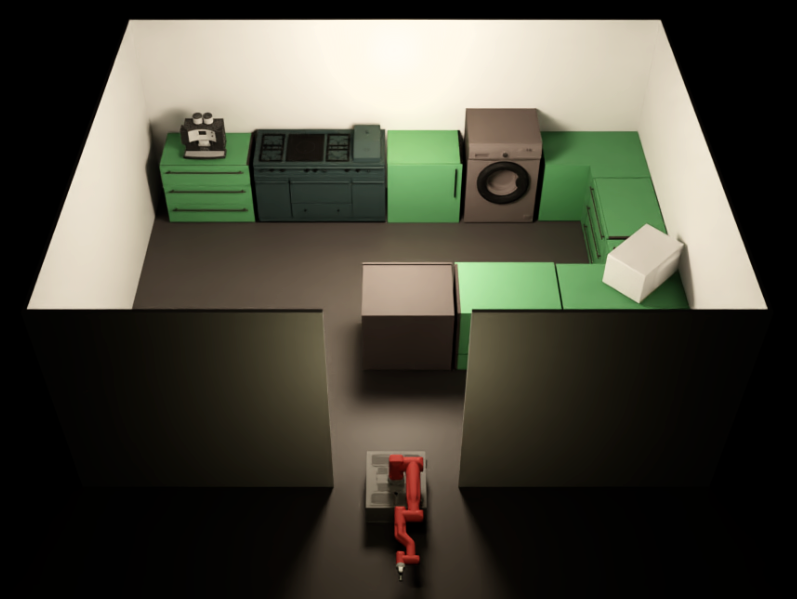}
    \includegraphics[width=0.225\linewidth]{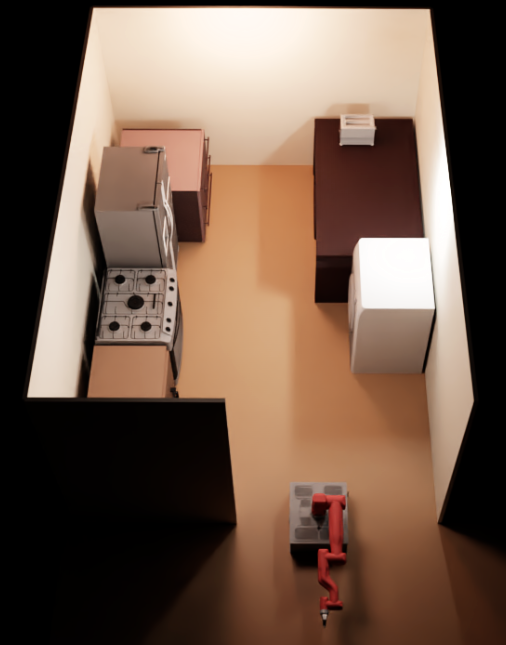}
    \caption{Different kitchen scenes designed in NVIDIA IsaacSim~\cite{nvidia2020omniverse}. The kitchens differ in architecture, free space for mobility, and articulated objects instances.}
    \label{fig:kitchen_layouts}
\end{figure}

We evaluate our method extensively in the high-fidelity simulator, NVIDIA Isaac Sim~\cite{nvidia2020omniverse}.
The simulator uses PhysX to provide stable, fast, realistic physics simulations and multi-GPU ray tracing for photo-realism.
We design three different kitchen layouts, as shown in~\figref{fig:kitchen_layouts}, using assets from PartNet-Mobility dataset~\cite{Xiang2020sapien}.
Each layout contains different everyday objects found in modern kitchens, such as tables, refrigerators, and microwaves. 
We add spring damping to all articulation instances to mimic realism. %

\subsection{Robotic platforms}
\label{sec:hardware_platform}

In simulation, we use the platform, \textit{Mabi-Mobile}~\cite{Pankert2020perceptivempc}. The robot has a differential drive base with four supporting castors and a 6-DoF manipulator. We attach the parallel jaw gripper by Franka Emika to the arm.
We mount four depth cameras on the base for 360$\si{\degree}$ Field-of-View and one RGB-D camera on the arm's shoulder. 
The ground-truth depth maps are clipped to $0.1~\si{\meter}-8.0~\si{\meter}$. The base cameras are used for online mapping, while the arm camera is used for perception.

We perform hardware tests on \emph{ANYmal-D} platform equipped with a 6-DoF torque controllable robotic arm, \emph{ALMA}~\cite{sleiman2021unified}. The arm has a Robotiq-2F85 parallel jaw gripper at the end-effector. We use the onboard Velodyne VLP-16 LiDAR for online mapping and the front-upper Intel Realsense camera on the base for perception.
The main control loop of WBC and state estimator runs at $400$~\si{\hertz}, while the MPC and the map update at $70$~\si{\hertz} and $15$~\si{\hertz} respectively.

\subsection{Object-centric planning using Articulation Estimator}

We consider the manipulation of washing machines, ovens/dishwashers, and drawers. For estimating the articulation parameters of these object categories, we use the pre-trained ANCSH networks in~\cite{Li2020ancsh}. In simulation, we use the object and handle's ground-truth segmentation masks. During hardware experiments, we obtain these using Mask-RCNN trained on LVIS Instance Segmentation Dataset~\cite{ wu2019detectron2}.

\subsection{Agent-centric Planning using MPC}
\label{sec:mpc_impl}

We implement the control module using OCS2 toolbox %
and use our adapted \textit{FIESTA} for 3D mapping at a resolution of $0.1$~\si{\meter}.
For the wheel-based mobile manipulator, we use a time horizon $t_h=4~\si{\s}$ and compute the optimal solution at $30~\si{\hertz}$. However, since the legged mobile manipulator has higher dimensions and more complex dynamics, we use a time-horizon $t_h=1~\si{\s}$ to solve the control problem faster ($70~\si{\hertz}$).
Instead of using a separate tracking module that follows the optimal whole-body trajectories, we evaluate the affine optimal policies with the latest state information and send the computed command to the motor controllers~\cite{Farshidian2017brealtimeplan}.

\subsection{Baseline agent-centric planners and Performance metrics}
\label{sec:baselines_impl}

In~\secref{sec:wbmpc}, we described the optimal control problem for whole body control (WBC) of mobile manipulation. Since our implementation uses SLQ method, we denote this method as \emph{SLQ-WBC}.

We compare SLQ-WBC to other agent-centric planners, which also use the robot model and represent the environment through an intermediate representation (occupancy map or ESDF). Since these approaches plan the base navigation to the object and object manipulation as separate phases, in all of them, we use RRT* to navigate in front of the object. During the interaction, the planners use different methods: i) fixed-base with arm-only control using SLQ-MPC~\cite{ruhr2012openingdoors,berenson2011tsr}, ii) WBC using IK~\cite{Burget2013wbmparticulated}, and iii) WBC using SLQ-MPC~\cite{Pankert2020perceptivempc}. We refer to these combinations as \emph{RRT* + SLQ-Arm}, \emph{RRT* + IK-WBC}, and \emph{RRT* + SLQ-WBC} respectively.

To evaluate the performances of the planners, we consider three metrics: the percentage of evaluated runs that are successful, the time taken to complete the task, and the path length traveled by the end-effector during successful executions. We consider the task completed if the joint position of the articulated object is more than a set threshold. These are defined at $60\%$, $65\%$, and $75\%$ of the maximum joint limit for ovens, drawers, and washing machines, respectively. 

\begin{table*}[t]
    \centering
    \caption{Comparison of SLQ-WBC against variants using RRT* for initial base navigation and a variant of the MPC for manipulation (SLQ-Arm, IK-WBC, and SLQ-WBC). We report the results averaged over 60 runs. The time taken and end-effector (EE) path length are reported for trajectories corresponding to task completion.}
    \label{tab:comparison_ik_slq}
    \begin{tabular}{c l c c c}
       \toprule
       \textbf{Category} & \textbf{Algorithm} & \textbf{Success Rate (in $\%$)} & \textbf{Time taken (in $\si{s}$)} & \textbf{EE Path length (in $\si{m}$)} \\ \midrule \midrule
        \multirow{4}{*}{\textbf{Drawer}} & RRT* + SLQ-Arm & $\color{Gray4}40.0$ & $\color{Gray4}16.409 \pm 3.353$ & $\color{Gray4}7.761 \pm 5.697$\\
        & RRT* + IK-WBC & $\color{Gray3}55.0$ & $\color{Gray3}17.843 \pm 1.571$ & $\color{Gray3}2.165 \pm 0.378$ \\
        & RRT* + SLQ-WBC & $\color{Gray1}\mathbf{95.0}$ & $\color{Gray2}15.177 \pm 0.729$ & $\color{Gray2}5.011 \pm 1.458$\\
        & SLQ-WBC & $\color{Gray2}93.3$ & $\color{Gray1}\mathbf{13.267 \pm 0.651}$ & $\color{Gray1}\mathbf{3.711 \pm 1.345}$\\ \midrule
        \multirow{4}{*}{\textbf{Oven}} & RRT* + SLQ-Arm & $\color{Gray4}0.0$ & -- & -- \\
        & RRT* + IK-WBC & $\color{Gray3}20.0$ & $\color{Gray3}11.523 \pm 4.562$ & $\color{Gray2}1.750 \pm 0.175$ \\
        & RRT* + SLQ-WBC & $\color{Gray1}\mathbf{90.0}$ & $\color{Gray2}9.121 \pm 0.659$ & $\color{Gray2}3.090 \pm 0.654$ \\
        & SLQ-WBC & $\color{Gray1}\mathbf{90.0}$ & $\color{Gray1}\mathbf{7.592 \pm 0.459}$ & $\color{Gray1}\mathbf{2.638 \pm 1.268}$ \\ \midrule
        \multirow{4}{*}{\textbf{Washing Machine}} & RRT* + SLQ-Arm & $\color{Gray4}0.0$ & -- & --\\
        & RRT* + IK-WBC & $\color{Gray3}46.6$ & $\color{Gray3}29.615 \pm 5.202$ & $\color{Gray2}4.243 \pm 0.916$ \\
        & RRT* + SLQ-WBC & $\color{Gray1}\mathbf{96.6}$ & $\color{Gray2}22.952 \pm 4.835$ & $\color{Gray3}6.213 \pm 4.071$\\
        & SLQ-WBC & $\color{Gray2}95.0$ & $\color{Gray1}\mathbf{18.747 \pm 2.452}$ & $\color{Gray1}\mathbf{3.907 \pm 0.739}$\\
        \midrule
        \midrule
        \multirow{4}{*}{\textbf{Overall}} & RRT* + SLQ-Arm & $\color{Gray4}13.3$ & $\color{Gray4}19.475  \pm 5.488$ & $\color{Gray4}7.267  \pm  4.831$\\
        & RRT* + IK-WBC & $\color{Gray3}40.5$ & $\color{Gray3}19.312  \pm 6.752$ & $\color{Gray3}3.854  \pm  1.397$ \\
        & RRT* + SLQ-WBC & $\color{Gray1}\mathbf{93.8}$ & $\color{Gray2}15.616 \pm 6.378$ & $\color{Gray2}4.559  \pm  2.855$\\
        & SLQ-WBC & $\color{Gray2}92.7$ & $\color{Gray1}\mathbf{14.152 \pm 4.935}$ & $\color{Gray1}\mathbf{3.644 \pm 1.301}$\\
        \bottomrule
    \end{tabular}

\end{table*}

\begin{figure*}
    \centering
    \includegraphics[width=\linewidth]{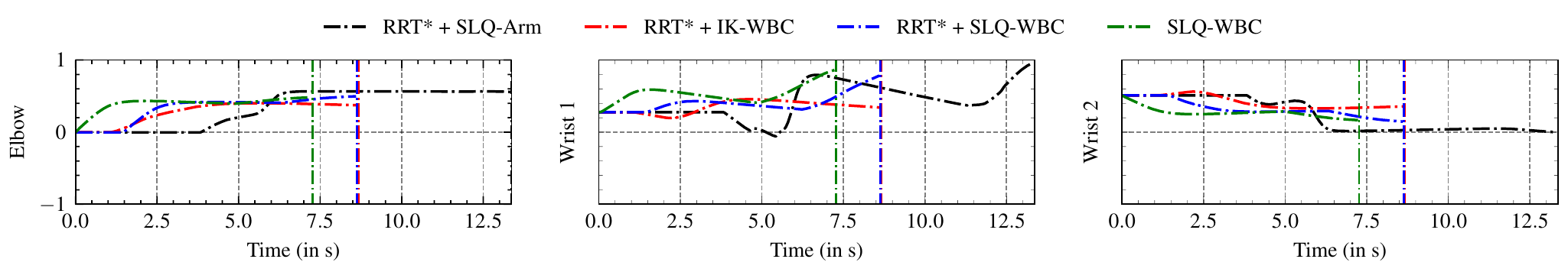}
    \caption{State evolution for last three arm joints of Mabi-Mobile during oven manipulation. The joint states are normalized w.r.t. their DoF limits. The vertical lines mark the end of the run, due to task completion, slippage, or hitting singularities.}
    \label{fig:oven_ik_fail}
\end{figure*}

\section{Results}
\label{sec:results}

\subsection{Benchmarking agent-centric planners}

We evaluate the agent-centric planners for mobile manipulators, described in~\secref{sec:baselines_impl}, on the simulated kitchen environments from~\secref{sec:sim_platform}. For each articulated object category in each room, we make 20 runs while spawning the robot randomly into the scene. Thus, a total of 60 runs are performed per articulated object category. The averaged performance metrics for the planners are reported in~\tabref{tab:comparison_ik_slq}.

The RRT* + SLQ-Arm baseline yields the most inferior performance across all three categories. Our experiments showed that this baseline failed to track the kinematic plan for ovens and washing machines manipulation. This is often because the robotic arm reaches its joint limits during the interaction, as shown in~\figref{fig:oven_ik_fail}. The joint limit of Wrist-1 in Mabi-Mobile is reached during oven manipulation by SLQ-Arm while other plans succeed in completing the task.

\begin{figure}
    \centering
    \begin{center}
        \begin{tikzonimage}[width=0.3\linewidth]{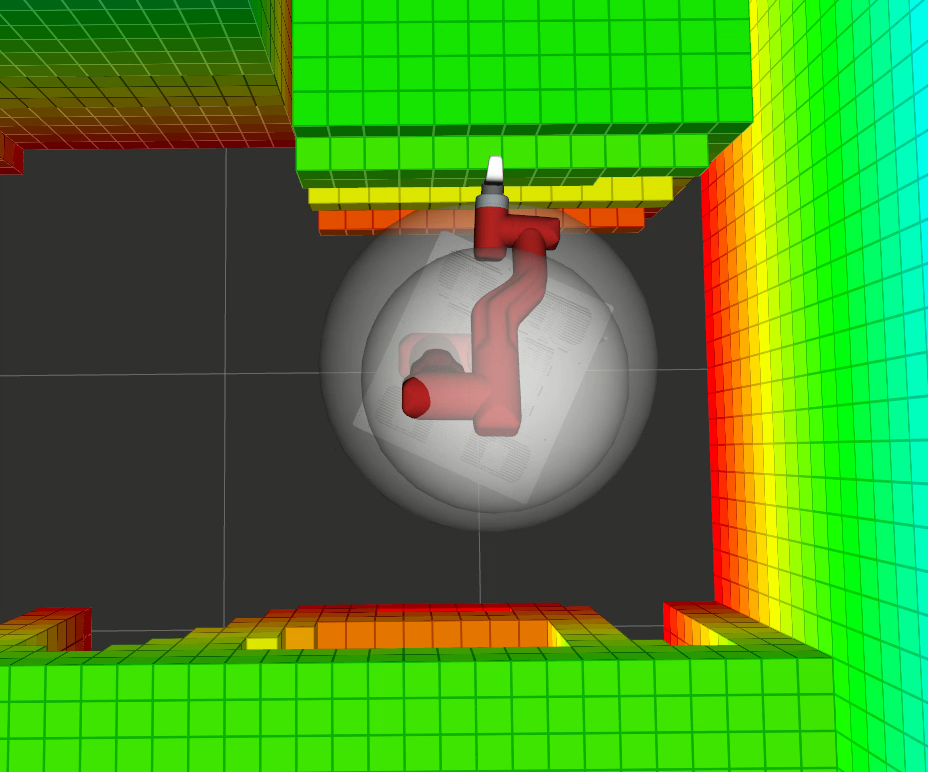}[image label]
        \node{a.1};
        \end{tikzonimage}
        \begin{tikzonimage}[width=0.3\linewidth]{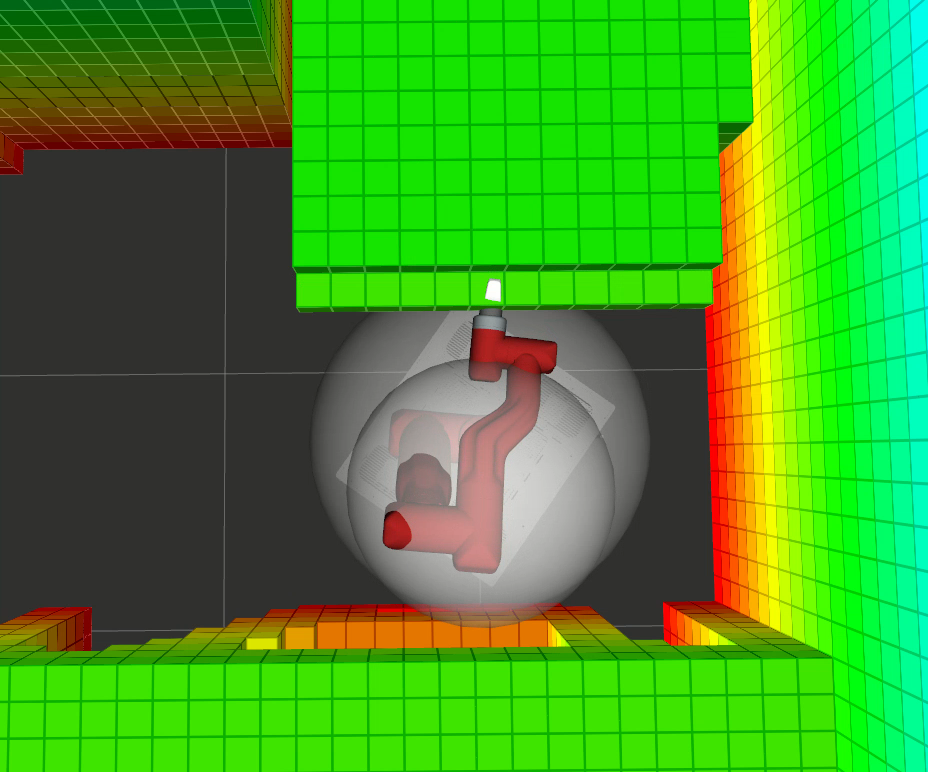}[image label]
            \node{a.2};
        \end{tikzonimage}
        \begin{tikzonimage}[width=0.3\linewidth]{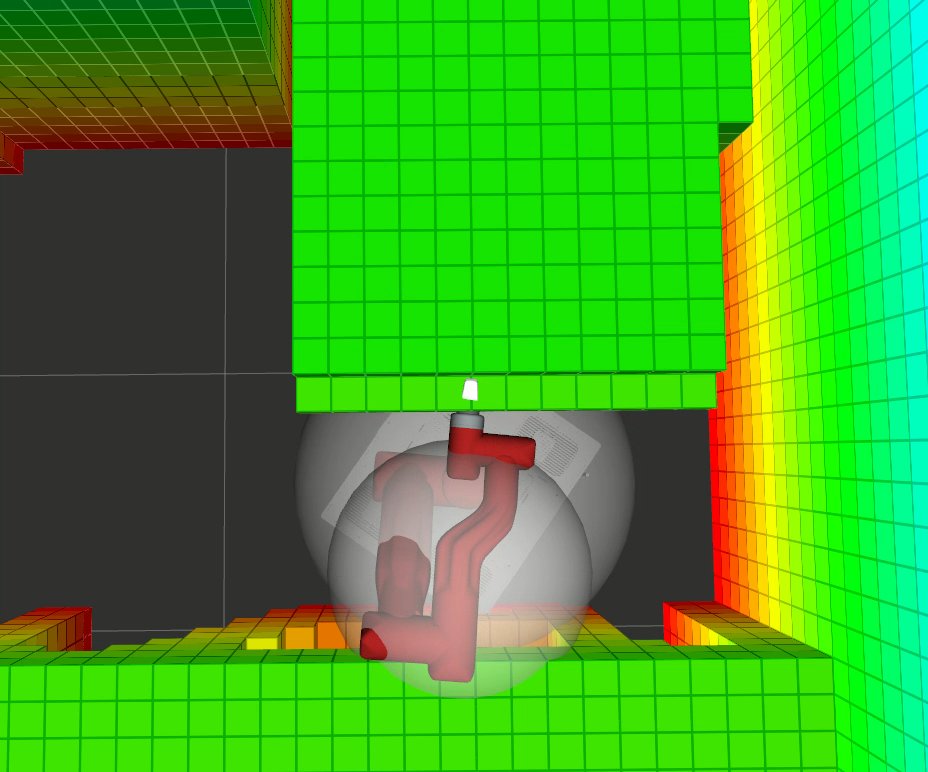}[image label]
            \node{a.3};
        \end{tikzonimage}
    \end{center}
    \hfill
    \\
    \vspace*{-20pt}
    \hfill
    \begin{center}
        \begin{tikzonimage}[width=0.3\linewidth]{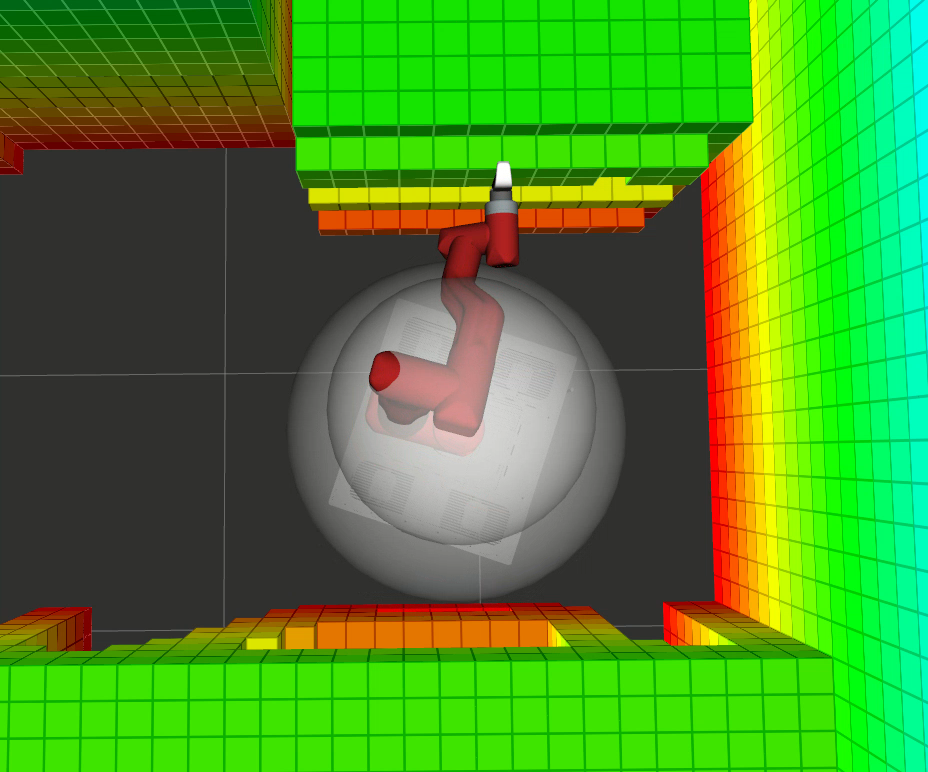}[image label]
        \node{b.1};
        \end{tikzonimage}
        \begin{tikzonimage}[width=0.3\linewidth]{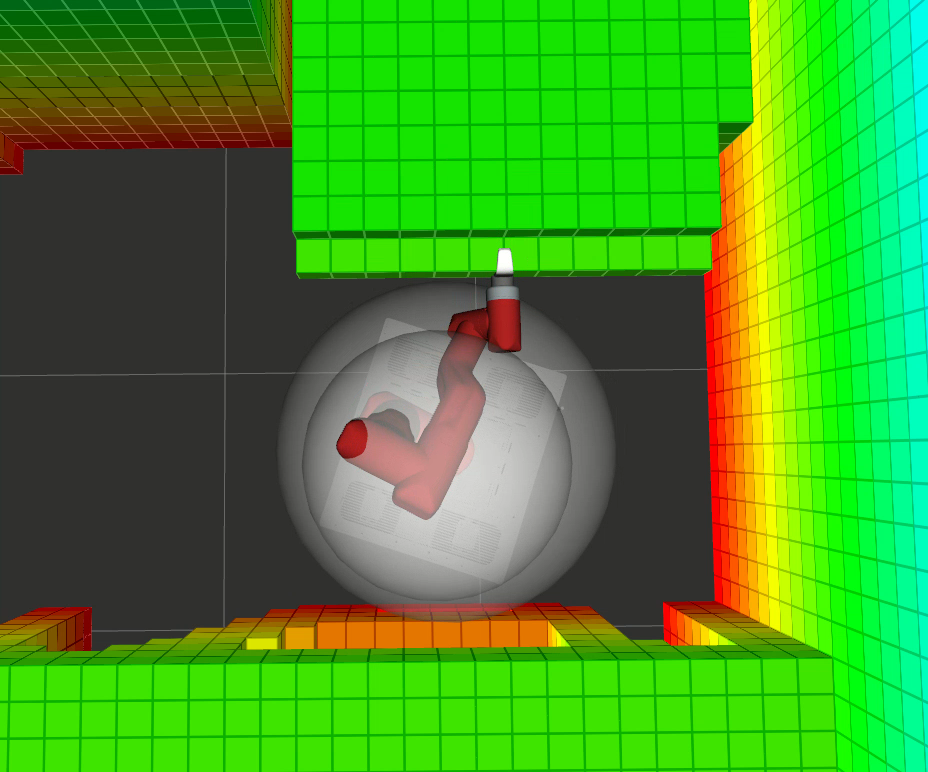}[image label]
            \node{b.2};
        \end{tikzonimage}
        \begin{tikzonimage}[width=0.3\linewidth]{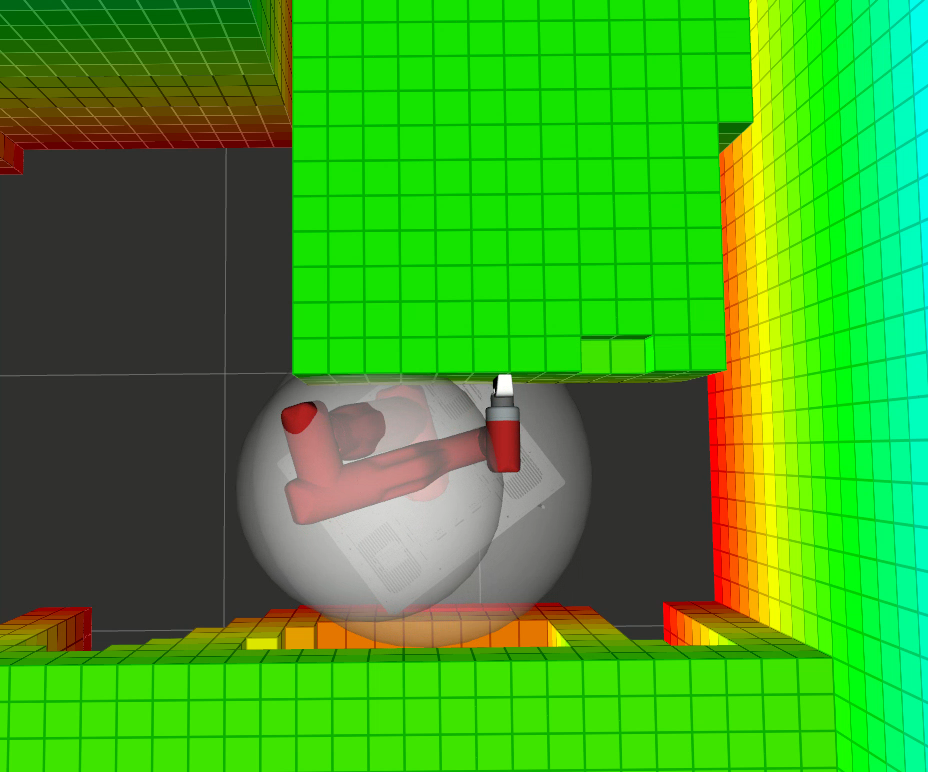}[image label]
            \node{b.3};
        \end{tikzonimage}
    \end{center}
    \hfill
    \\
    \vspace{-4pt}
    \caption{
    Reactivity using (a) IK and (b) MPC. A longer preview horizon helps the robot to plan ahead and avoid collisions.
    }
    \label{fig:ik_vs_slq_rviz}
\end{figure}

A longer horizon helps plan the base and arm coordination better for collision avoidance. However, it comes at the cost of computation time for the MPC.~\figref{fig:ik_vs_slq_rviz} shows an example from drawer manipulation where the short look-ahead of IK-WBC drives the base to a pose from which avoiding environment collision is hard. The difference in the performance of RRT* + IK-WBC and RRT* + SLQ-WBC quantitatively establishes the importance of a look-ahead for planning. Compared to RRT* + IK-WBC, SLQ-WBC improves the overall success rate by $134\%$ and reduces average task completion time by $26.5\%$.

Failure cases for the SLQ-WBC attribute to: i) incorrect articulated properties estimation due to bad initial viewpoints (severe occlusions), and ii) convergence to sub-optimal solutions where the gradient in favor of achieving the task competes with the one for preventing possible collisions. By incorporating a global planner to escape local optima, the RRT* + SLQ-WBC circumvents the issue; thus has a higher success rate than SLQ-WBC alone.
However, since the RRT* + SLQ-WBC first performs base navigation and then whole-body control during interaction, it is less efficient in executing the task than SLQ-WBC. 
We believe a better approach for escaping local optima in MPC optimization is utilizing the global plan to initialize the MPC solver.

\begin{figure*}
    \centering
    \hfill
    \begin{center}
        \begin{tikzonimage}[width=0.19\linewidth]{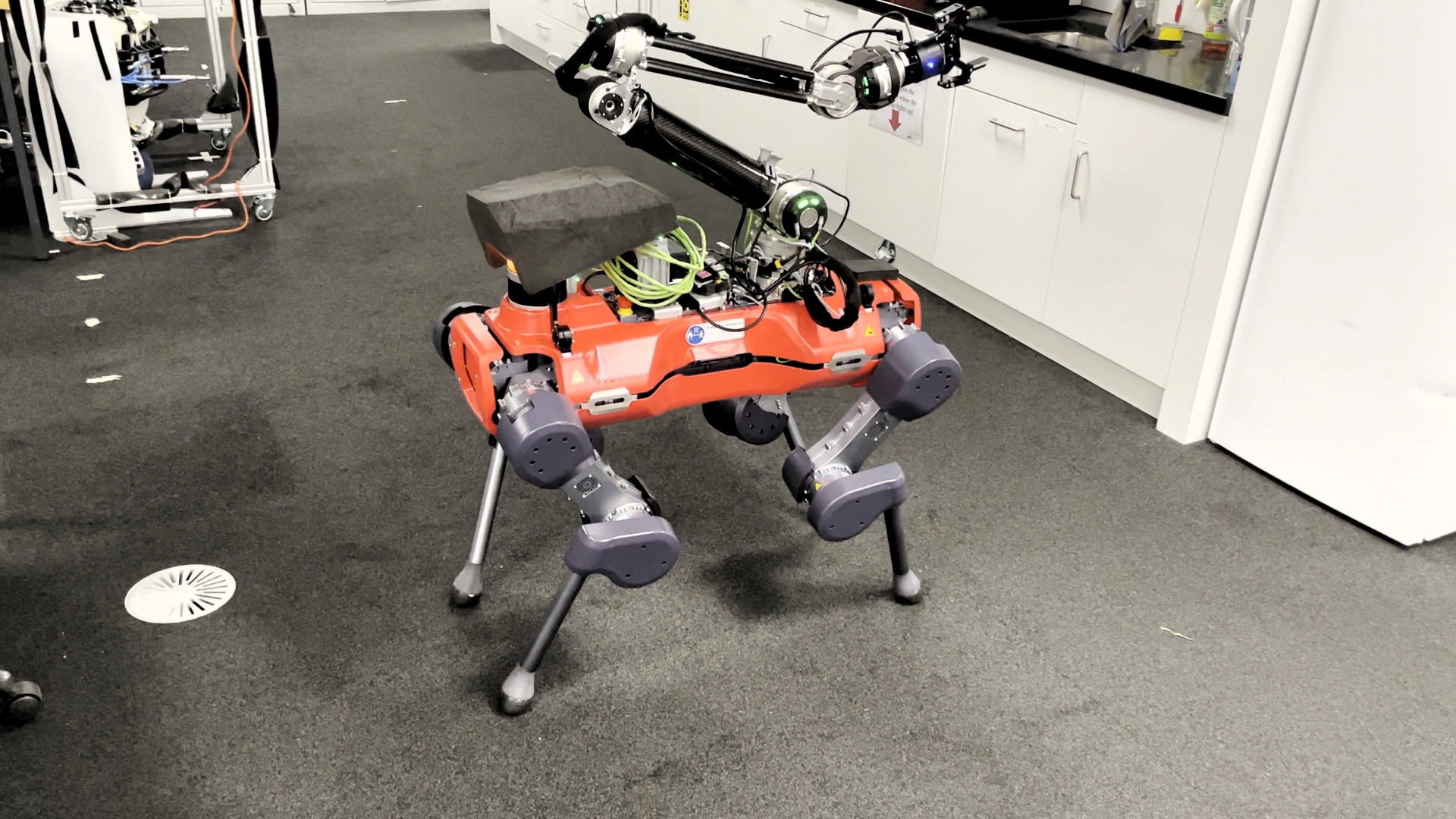}[image label]
        \node{a.1};
        \end{tikzonimage}
        \begin{tikzonimage}[width=0.19\linewidth]{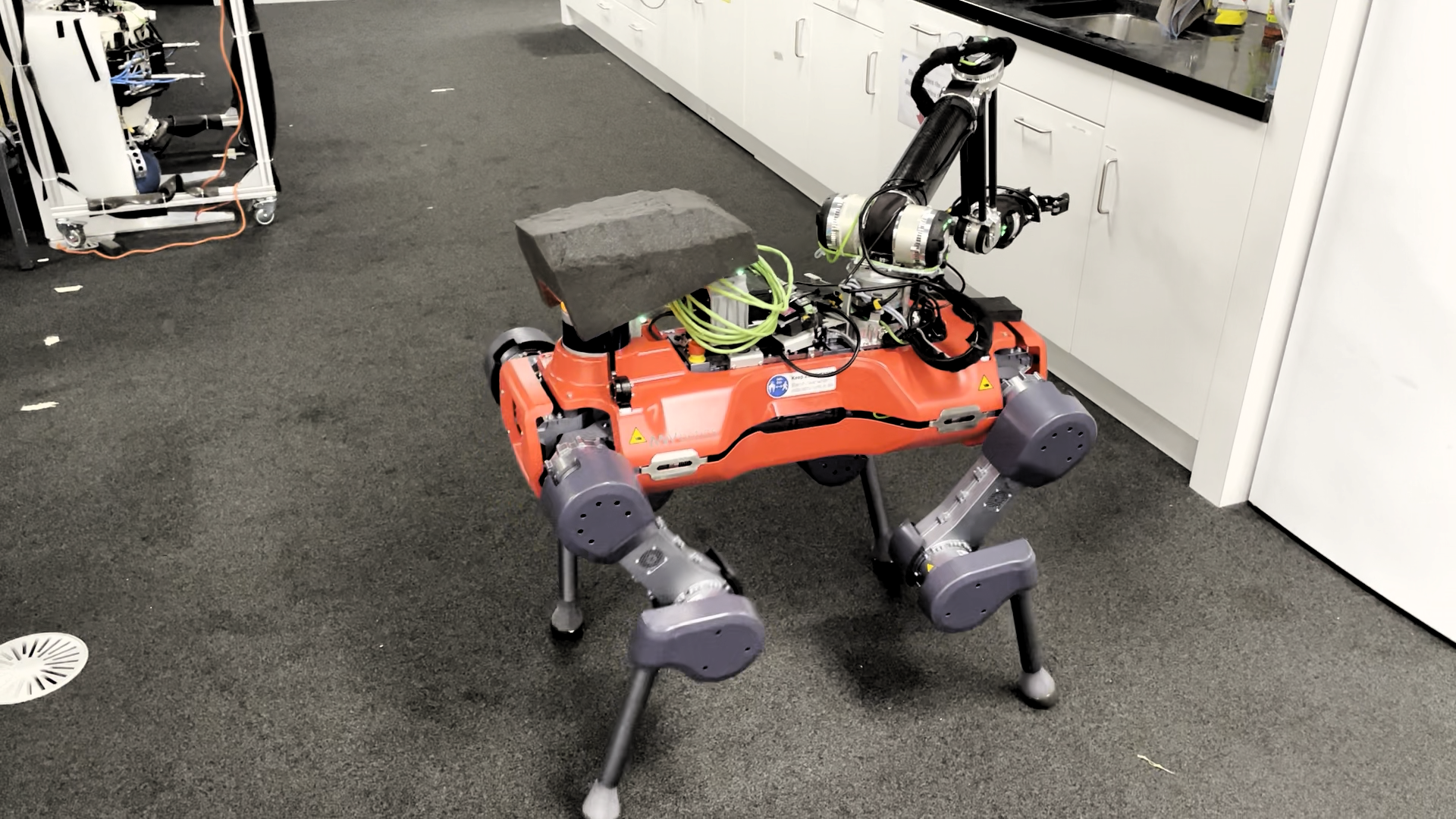}[image label]
        \node{a.2};
        \end{tikzonimage}
        \begin{tikzonimage}[width=0.19\linewidth]{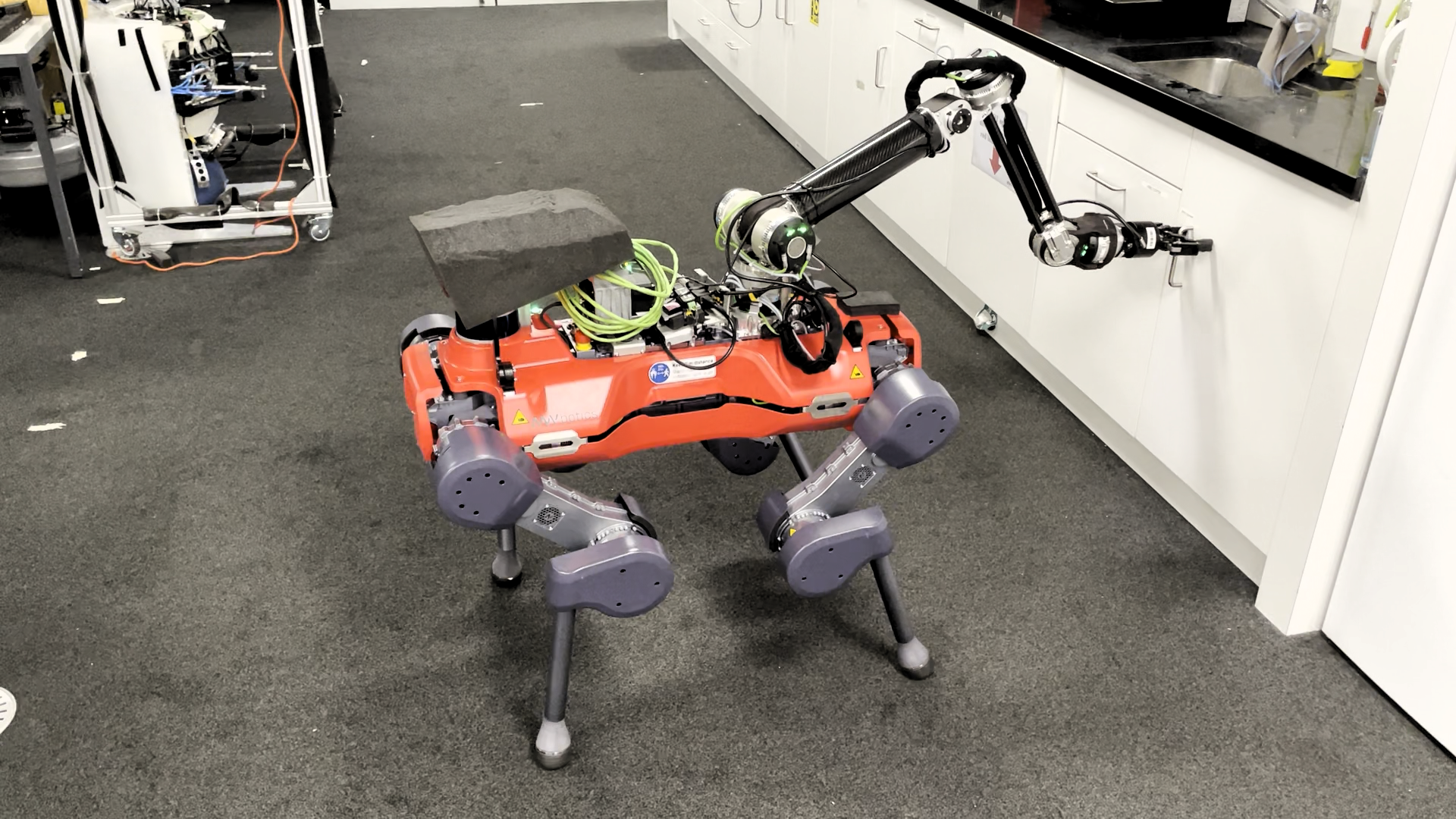}[image label]
        \node{a.3};
        \end{tikzonimage}
        \begin{tikzonimage}[width=0.19\linewidth]{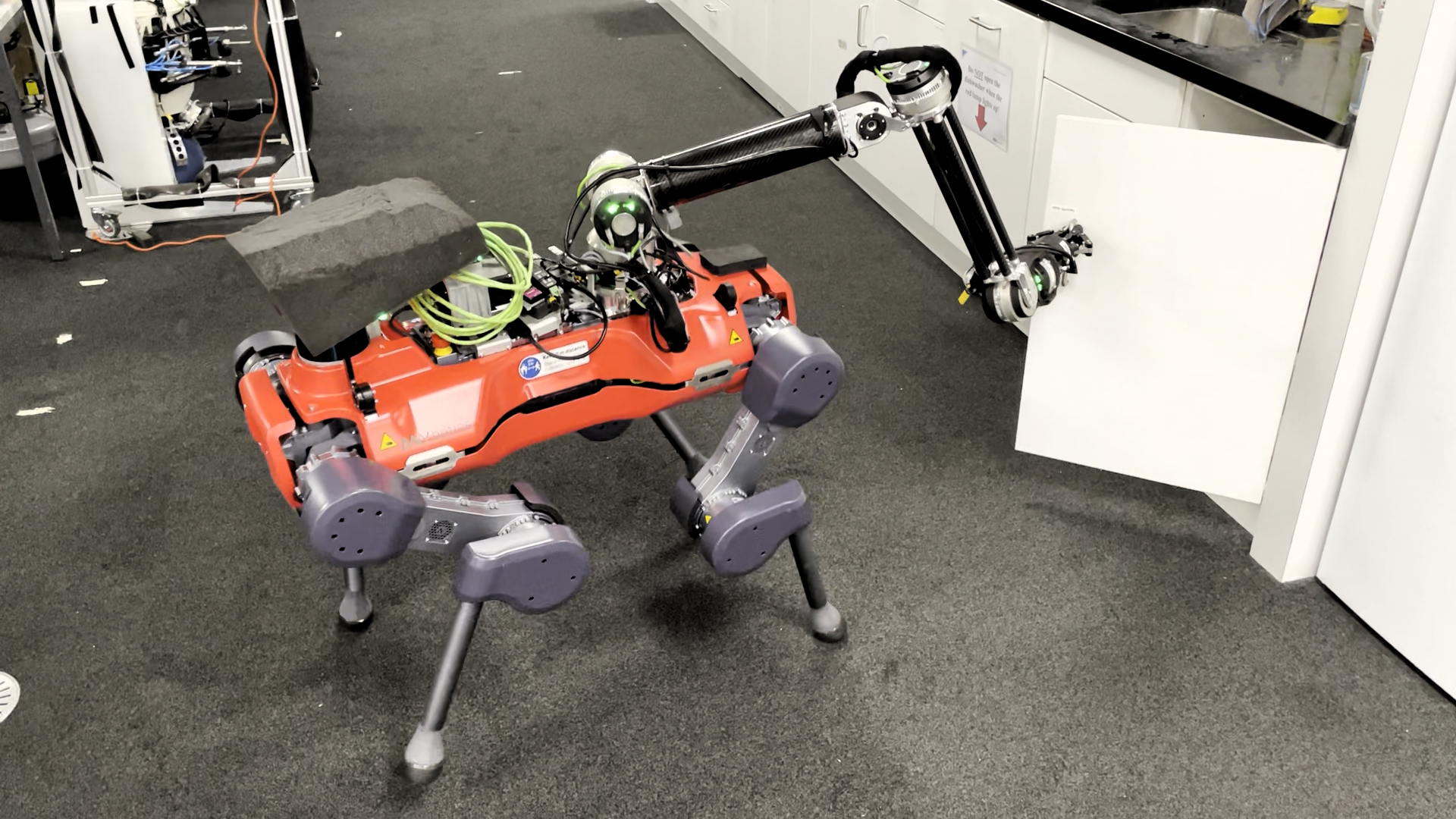}[image label]
        \node{a.4};
        \end{tikzonimage}
        \begin{tikzonimage}[width=0.19\linewidth]{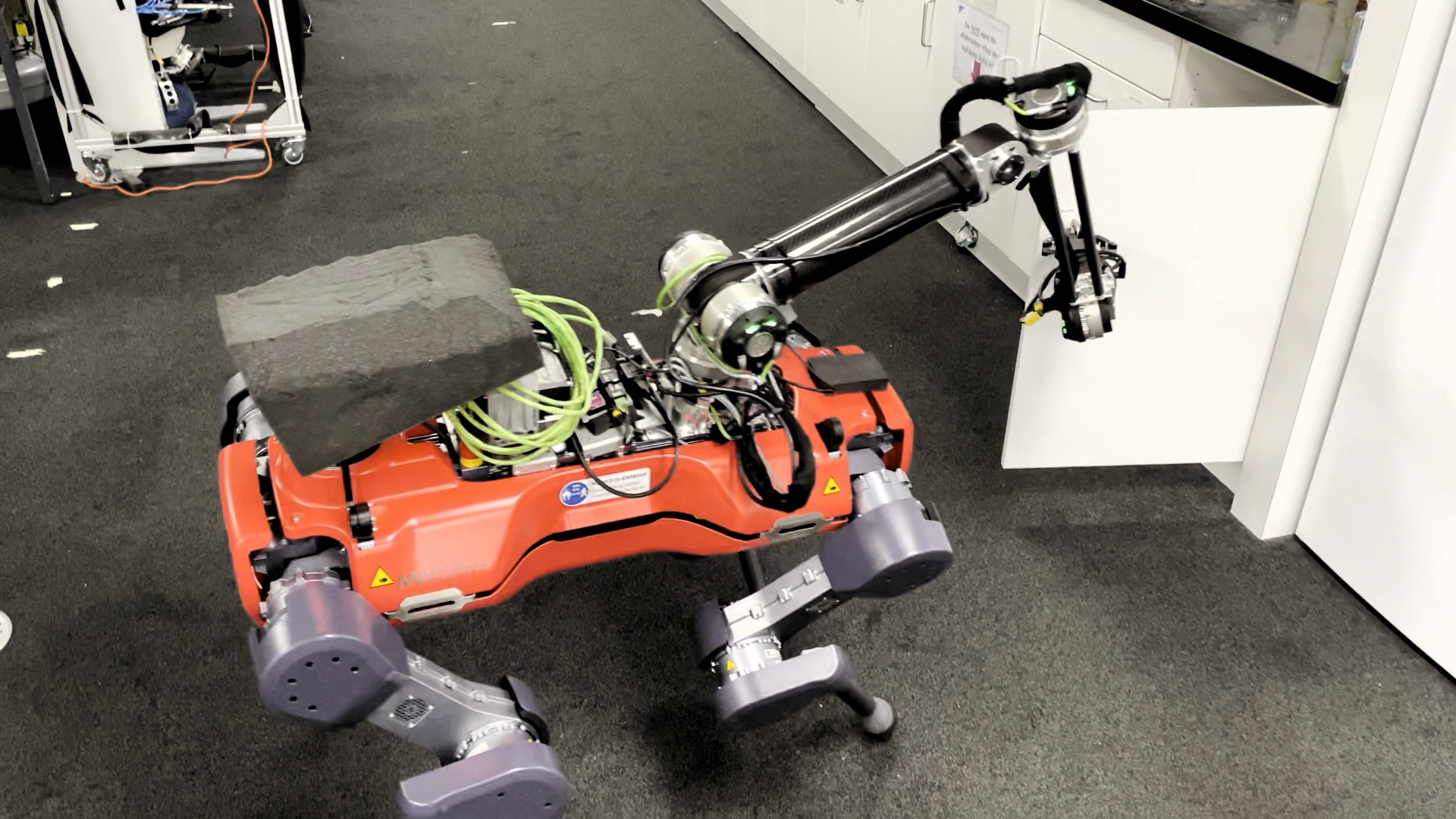}[image label]
        \node{a.5};
        \end{tikzonimage}
    \end{center}
    \hfill
    \begin{center}
        \begin{tikzonimage}[width=0.19\linewidth]{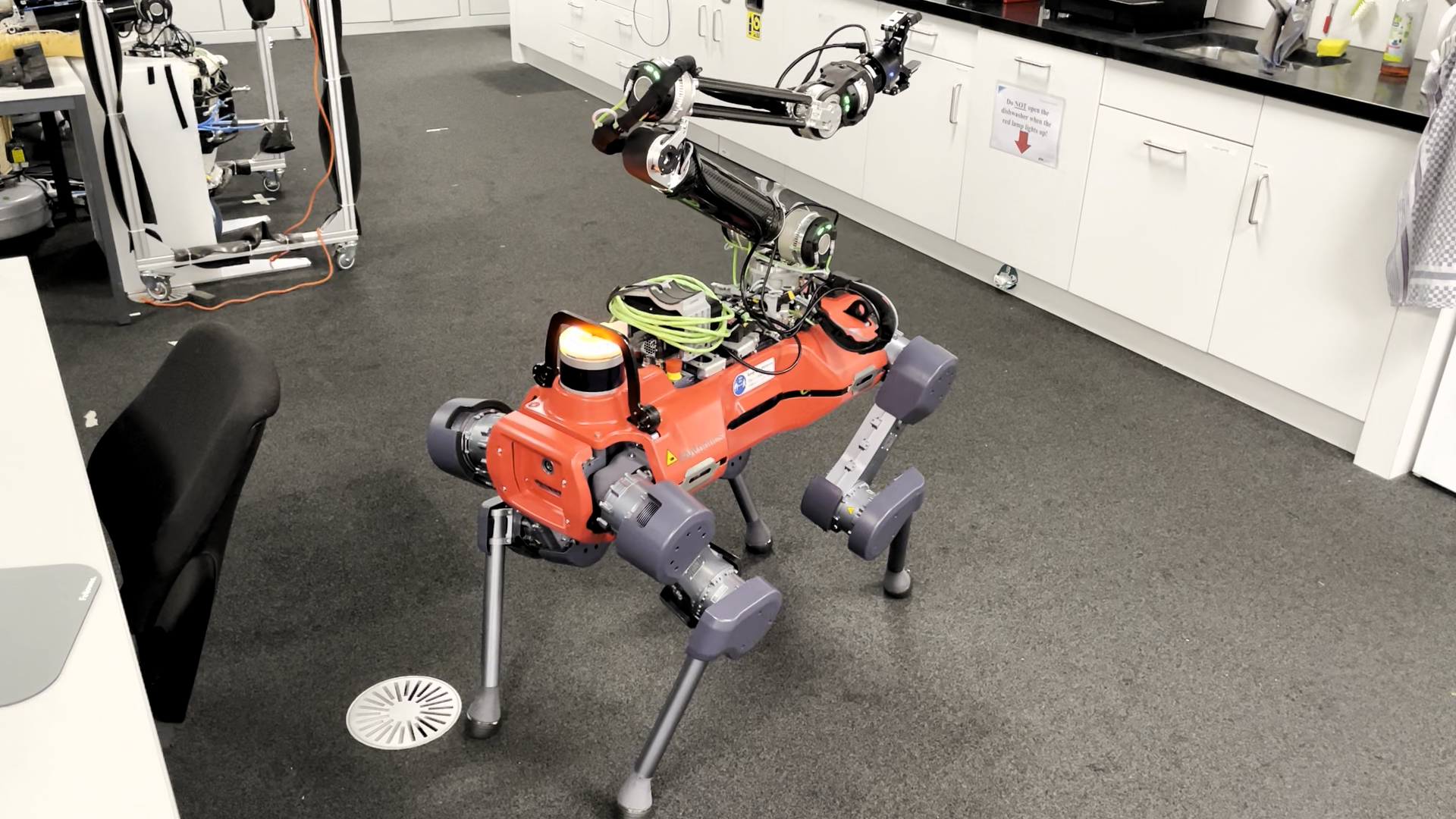}[image label]
        \node{b.1};
        \end{tikzonimage}
        \begin{tikzonimage}[width=0.19\linewidth]{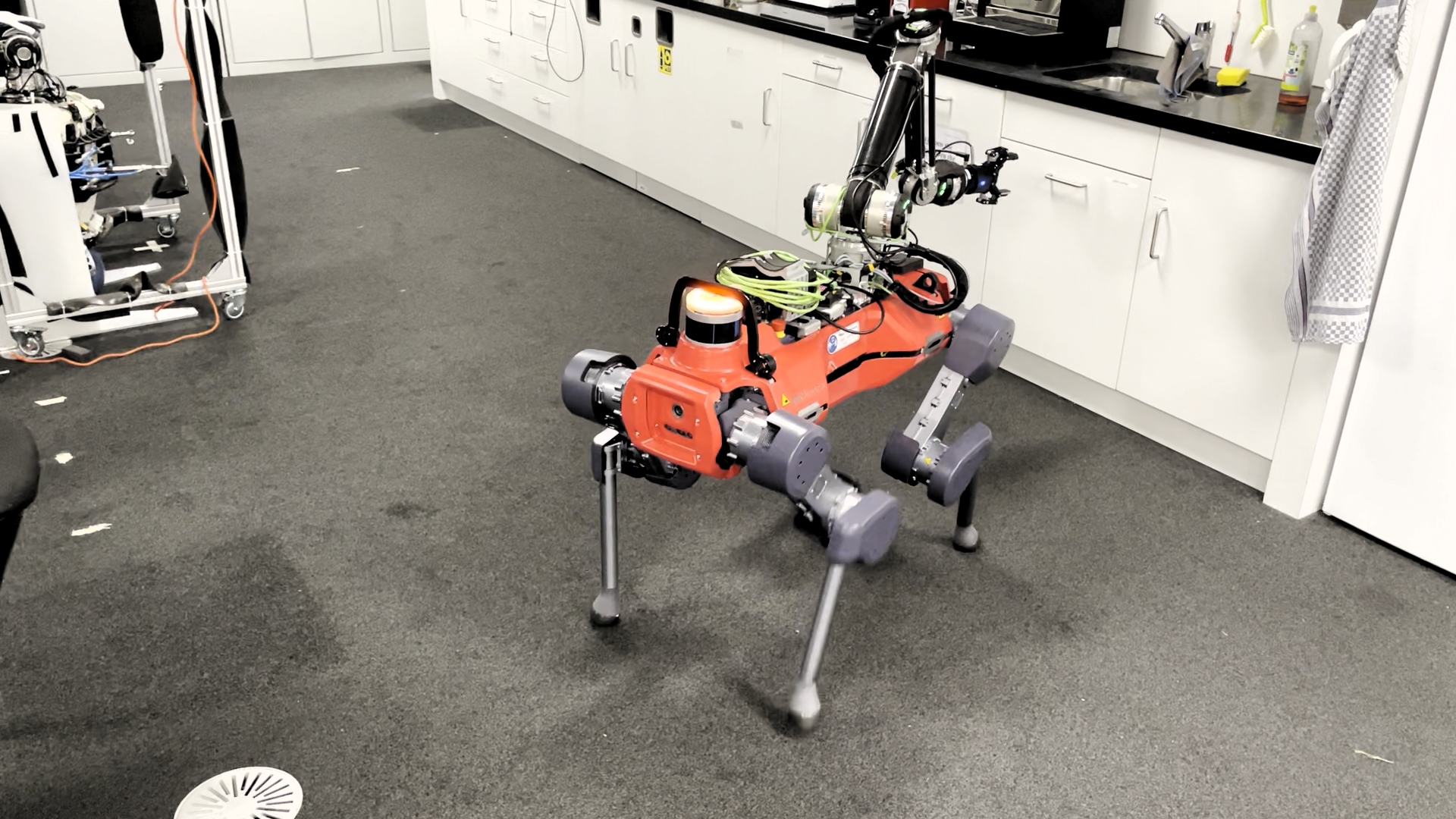}[image label]
        \node{b.2};
        \end{tikzonimage}
        \begin{tikzonimage}[width=0.19\linewidth]{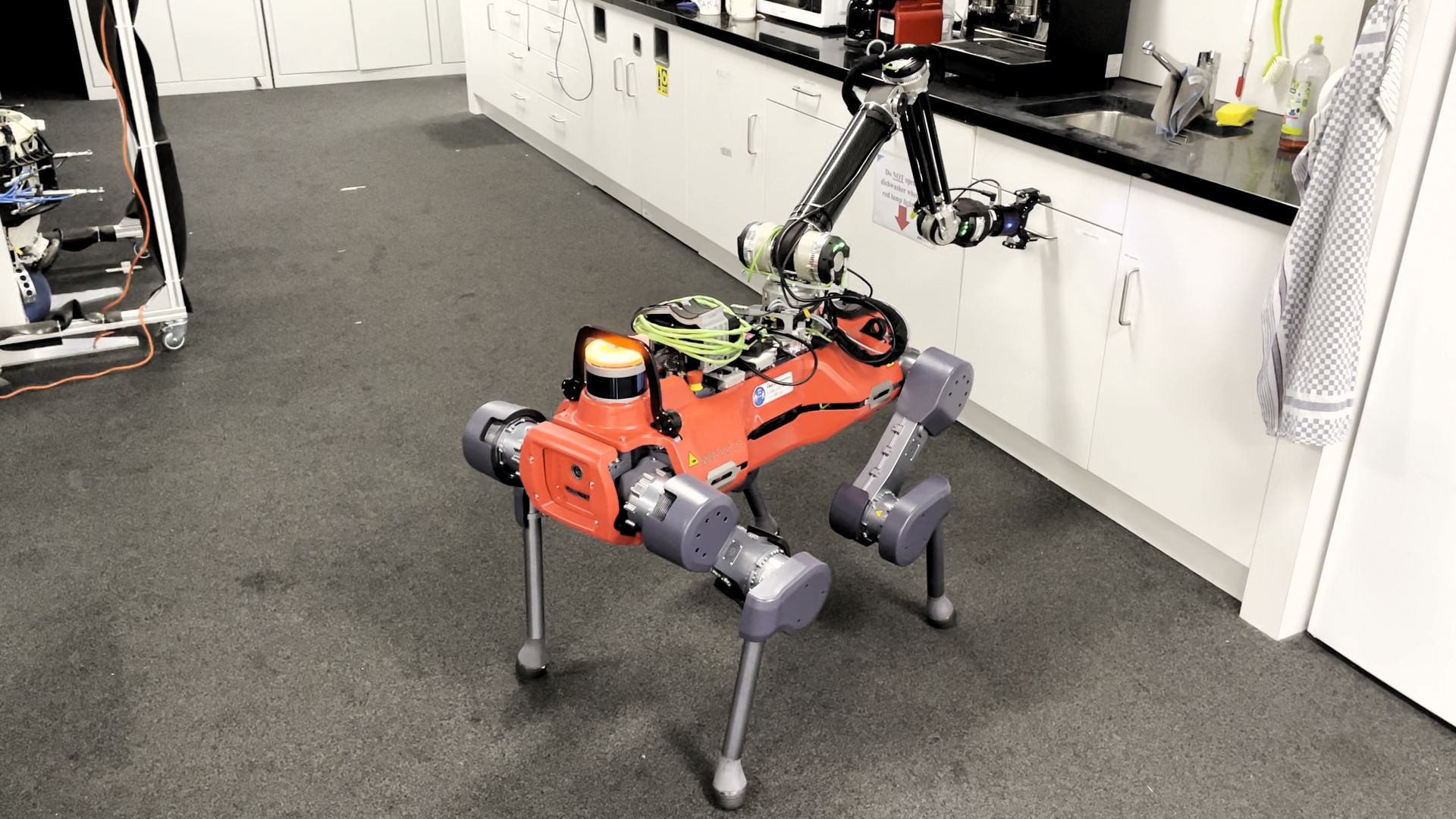}[image label]
        \node{b.3};
        \end{tikzonimage}
        \begin{tikzonimage}[width=0.19\linewidth]{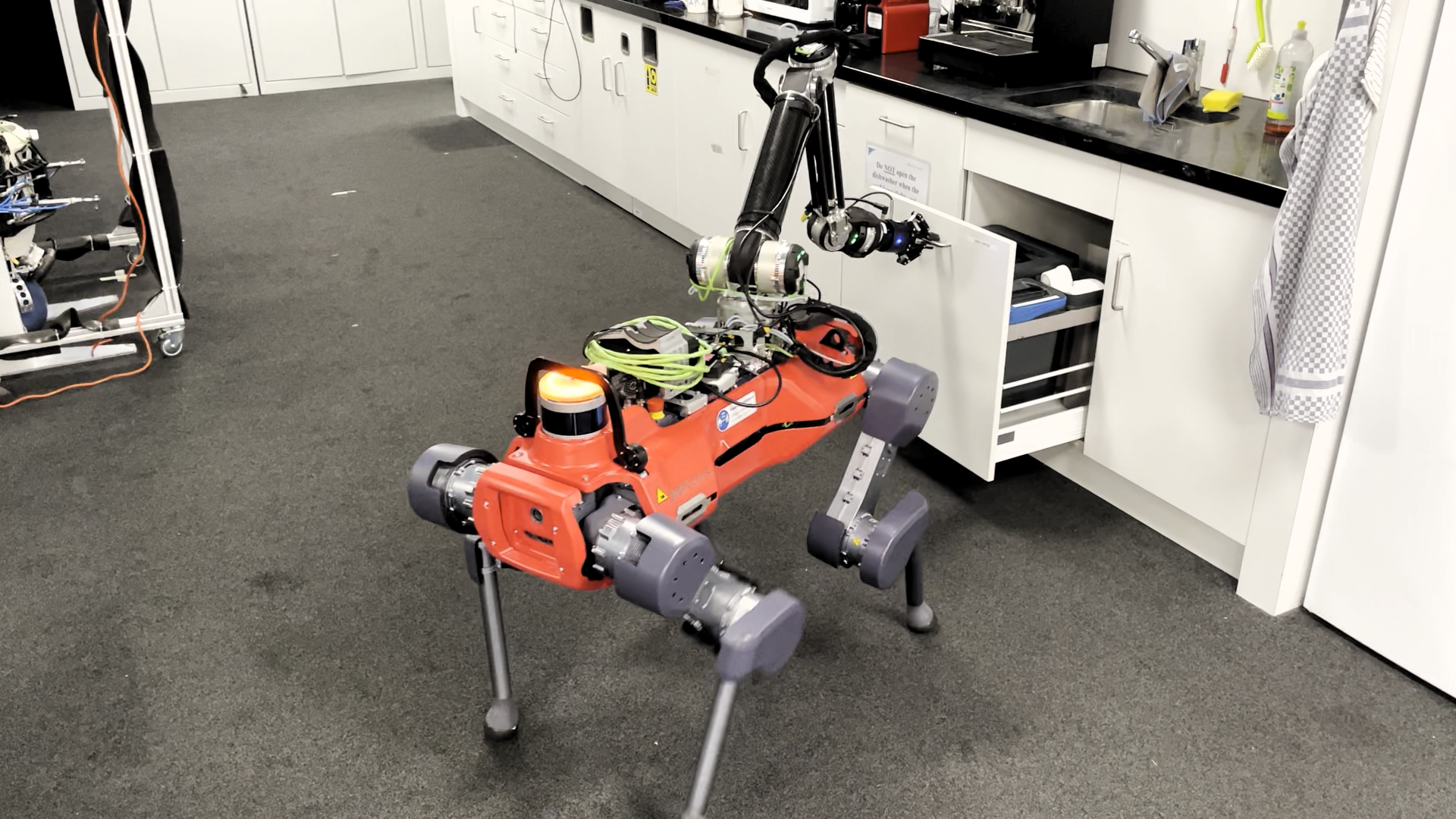}[image label]
        \node{b.4};
        \end{tikzonimage}
        \begin{tikzonimage}[width=0.19\linewidth]{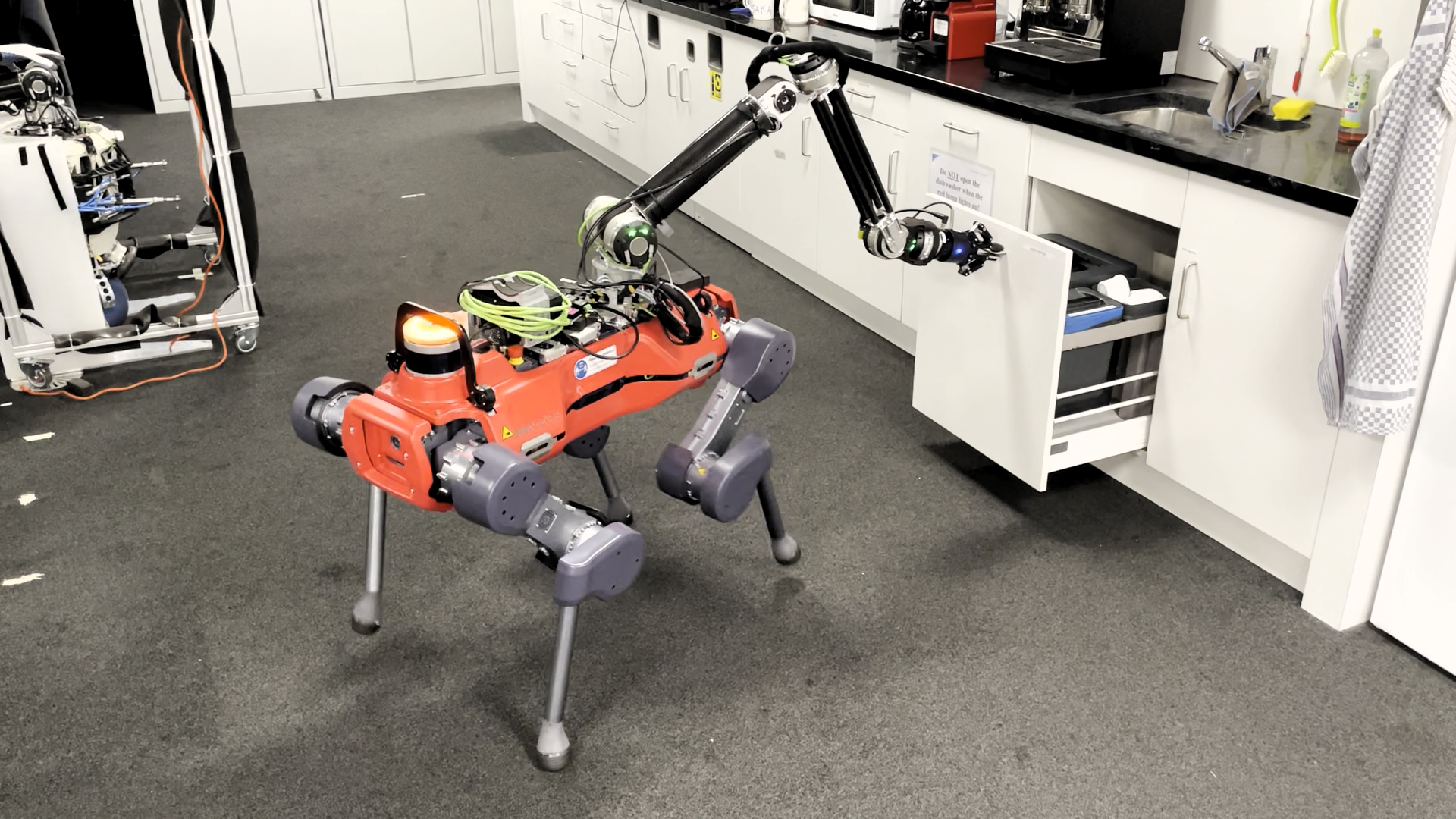}[image label]
        \node{b.5};
        \end{tikzonimage}
    \end{center}
    \hfill
    \vspace{-12pt}
    \caption{Legged mobile manipulation of articulated objects in our kitchen testbed: (a) Drawer, (b) Cabinet. Throughout the interaction, we set the robot gait schedule to trot. Only while grasping the handle, the robot enters stance mode.}
    \label{fig:alma_hardware}
\end{figure*}

\subsection{Deployment in a real kitchen}

On hardware, we found that ANCSH worked well when the objects were in isolation. However, it performed poorly in our kitchen testbed, where things are tightly packed, and only the moving links of the objects are visible.
As a workaround for hardware, we employ heuristics based on the object-category and RANSAC-based plane-fitting to compute the articulation properties.

We deploy the resulting system on \emph{ALMA} to test the manipulation of various articulated objects, as shown~\figref{fig:alma_hardware}. %
During the approach and manipulation phase, we set the robot's gait schedule to trot. However, we observed that the trotting motion led to a shaky end-effector, making grasps imprecise. Thus, when the robot is in vicinity to the handle, we temporarily set it to stance mode.

\begin{figure}[!t]
    \centering
    \includegraphics[width=\linewidth]{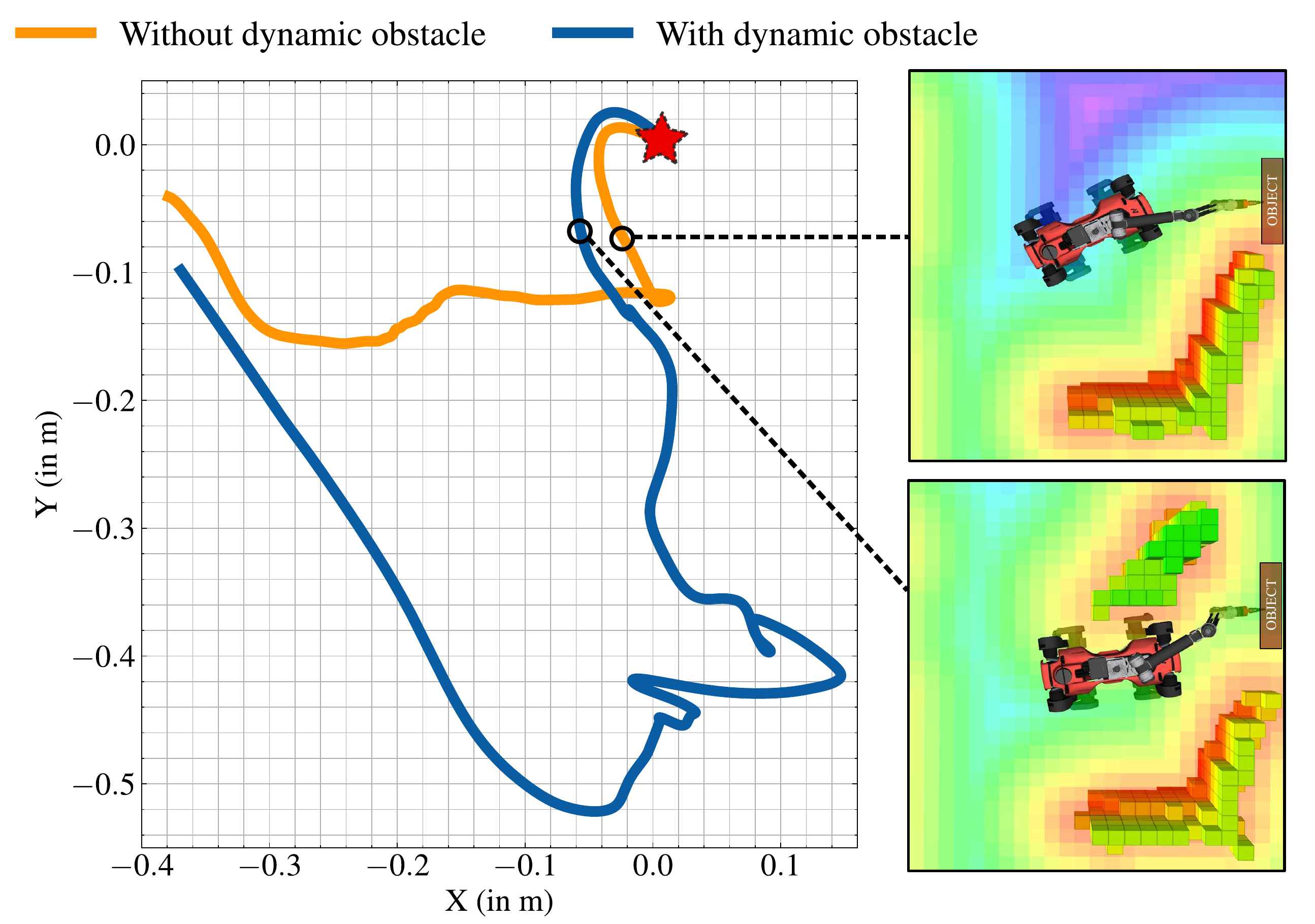}
    \caption{\textbf{Left:} Plot showing 2D motion of the legged base with and without dynamic obstacle in the scene. The red star marks the base position at the start of manipulation. \textbf{Right:} Visualization of the occupancy map and ESDF created online at the marked locations.}
    \label{fig:dynamic_alma}
\end{figure}

\subsection{Dynamic-collision avoidance during object interaction}
\begin{figure*}[!t]
    \centering
    \hfill
    \begin{center}
        \begin{tikzonimage}[width=0.19\linewidth,trim=0 50 0 0, clip]{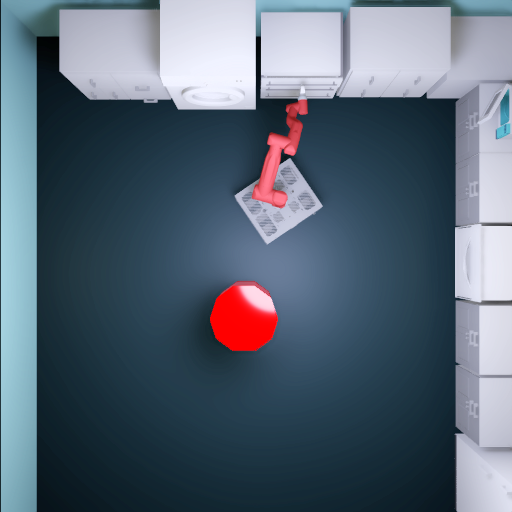}[image label]
        \node{1};
        \end{tikzonimage}
        \begin{tikzonimage}[width=0.19\linewidth,trim=0 50 0 0, clip]{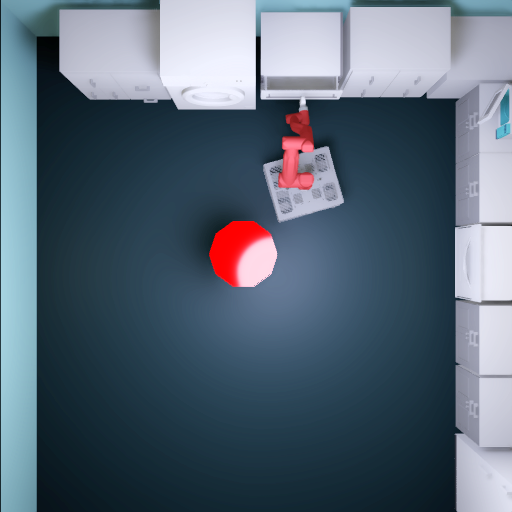}[image label]
            \node{2};
        \end{tikzonimage}
        \begin{tikzonimage}[width=0.19\linewidth,trim=0 50 0 0, clip]{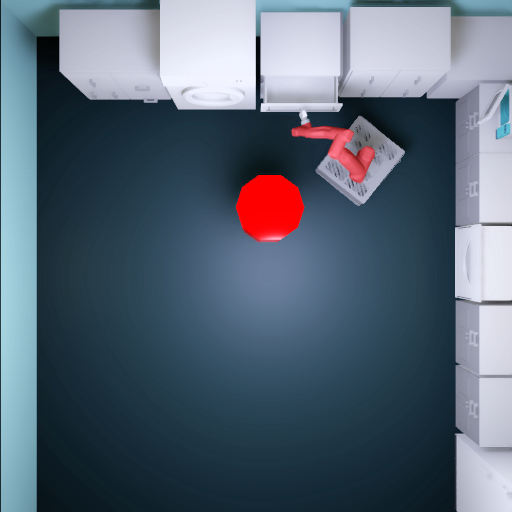}[image label]
            \node{3};
        \end{tikzonimage}
        \begin{tikzonimage}[width=0.19\linewidth,trim=0 50 0 0, clip]{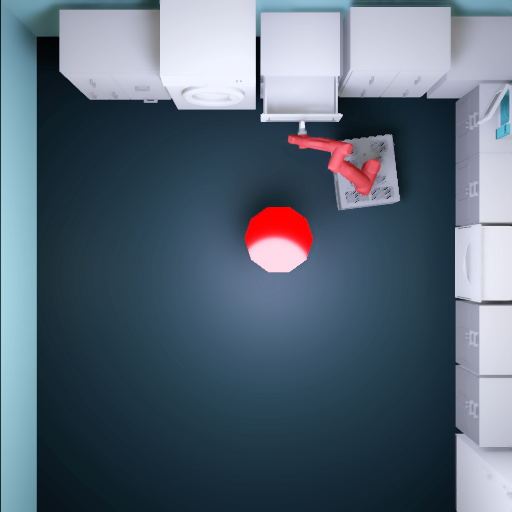}[image label]
            \node{4};
        \end{tikzonimage}
        \begin{tikzonimage}[width=0.19\linewidth,trim=0 50 0 0, clip]{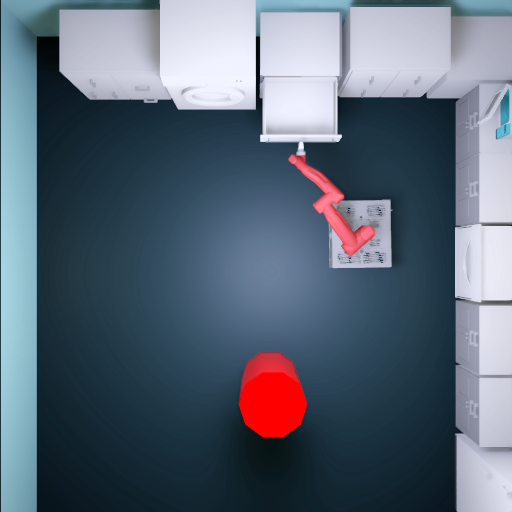}[image label]
            \node{5};
        \end{tikzonimage}
    \end{center}
    \hfill
    \\
    \vspace{-3pt}
    \caption{Drawer manipulation in a dynamic scene where the red obstacle moves at 0.5 \si[per-mode=symbol]{\meter\per\second}. The reactive agent-centric planner, which updates ESDF online, ensures tracking of the given object-centric plan while preventing collisions.}
    \label{fig:dynamic_awesome_manip}
\end{figure*}

We first show the online collision avoidance during object manipulation in a controlled simulated setting (\figref{fig:dynamic_awesome_manip}). Here the object moves to randomly sampled goals at a speed of 0.5 \si[per-mode=symbol]{\meter\per\second}. The agent-centric planner, which updates the ESDF map online and uses it in MPC, ensures tight coordination of the base and the arm to prevent any collisions\footnote{For videos, please check: \href{https://www.pair.toronto.edu/articulated-mm/}{www.pair.toronto.edu/articulated-mm/}
}.

We also test dynamic collision avoidance on hardware. In~\figref{fig:dynamic_alma}, we plot the base motion during drawer manipulation for two scenarios: with and without a dynamic obstacle. In both these scenarios, the robot starts from the same location. The dynamic obstacle moves towards the robot during the manipulation phase. Without a dynamic obstacle in the scene, the base back fairly straight. However, when the dynamic obstacle is present, it starts moving away from it and uses the arm more to ensure tracking of the interaction trajectory. Once the dynamic obstacle moves away, the base retreats to a configuration that reduces the joint limits and self-collision avoidance penalties.

Since our implemented MPC is only \textit{reactive}, it fails to plan ahead for obstacles moving at high speeds. A possible solution is using a motion model for moving obstacles to predict their future locations. We leave this for future work.

\begin{figure}[!t]
    \centering
    \includegraphics[width=0.925\linewidth]{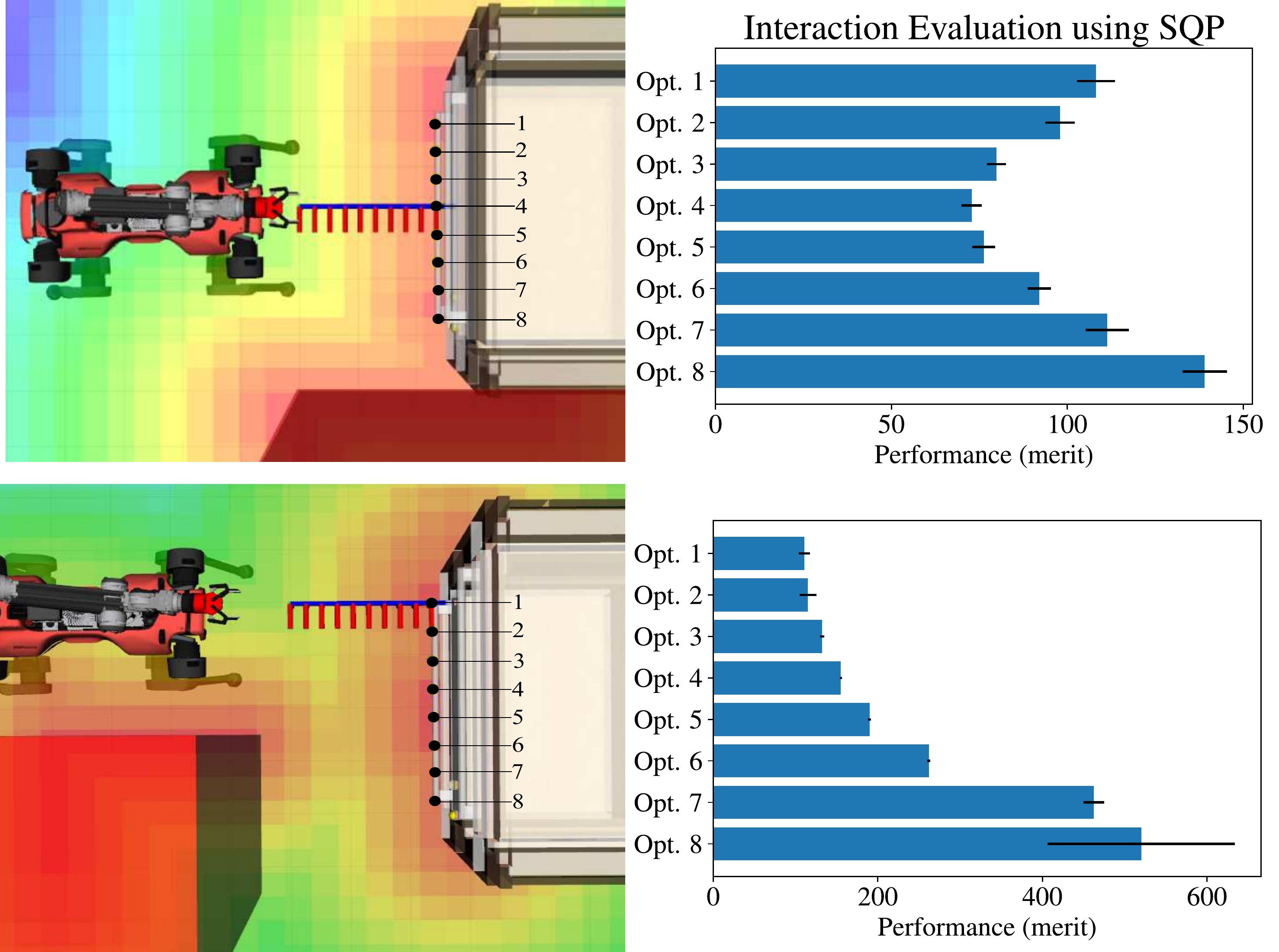}
    \vspace{2pt}
    \caption{Interaction options evaluation with obstacle beside and in-front of the object. A lower merit implies a safer option. We use SQP to approximately solve~\eqref{eq:op} for a fixed number of iterations. The bar plot shows the mean and standard-deviation over five runs.}
    \label{fig:options_eval}
\end{figure}

\subsection{Decision-making between interaction options}

In the previous results, we focused mainly on motion generation by the agent-centric planner and considered the most common interaction option, i.e., by grasping the object's handle at its center. However, this option may not always be the most suitable due to obstacles. The following experiment presents a preliminary result on utilizing the agent-centric planner to choose between multiple object-centric plans.

We consider the manipulation of a drawer with a long handle. By uniformly discretizing the grasp locations on the handle, we generate interaction plans. The agent-centric planner evaluates the \emph{merit function} for these plans by using the formulation~\eqref{eq:op} with a long planning horizon $t_h=10\si{s}$. The merit function~\cite{wright1999numerical} combines the task objective with measures of constraints violation. A lower merit implies a safer interaction plan. 

For two different static scenes, we observe that the same interaction plans yield different merits, as reported in~\figref{fig:options_eval}. Plans that are closer to the obstacle have higher scores due to the collision avoidance penalty (\emph{environment cost}). When the collision possibility is low, other factors such as the input cost and joint constraints play a role in decision-making (\emph{embodiment cost}). Thus, by combining these two factors, the proposed formulation for the agent-centric planner can also help in deciding between interaction plans.

\section{Conclusion}
\label{sec:conclusion}

We developed a two-stage hierarchy comprising object-centric and agent-centric planning for autonomous mobile manipulation of large articulated objects.
For such a decomposition, we discussed the properties desired from the agent-centric planner to allow decision-making and tracking of object-centric plans in unknown static and dynamic scenes.
We formulated these properties into an optimal control problem and showed the resulting system's efficacy in simulated and real-world scenarios for wheeled-base and legged mobile manipulators.
To perform online collision avoidance, we adapted FIESTA to update robot-centric SDF and a globally-consistent occupancy map.
In simulation, we benchmarked our approach to other choices for agent-centric planners, which use sampling or IK for planning. 
We showed that our proposed system has a higher success rate and a lower execution time for manipulating articulated objects. 

We open-source our implementation for the MPC-based agent-centric planner as a part of OCS2 Toolbox\footnote{Implementation available at: \href{https://github.com/leggedrobotics/ocs2}{https://github.com/leggedrobotics/ocs2}}. Our implementation is generic and works for any fixed arm or wheel-based mobile manipulator given its URDF.

Since we primarily investigated a suitable agent-centric planner, a natural next step is combining it with a better object-centric planner to generate prehensile and non-prehensile interactions. Additionally, while our result on deciding from a set of candidate object-centric plans is promising, we assumed that all grasp options are equal and stable. However, that may not always hold, and we need to also account for the confidence in the interaction itself (grasping handle \textit{vs.} edges). Applying data-centric methods is a promising direction to overcome these limitations.

\bibliographystyle{bibtex/myIEEEtran} 
\bibliography{bibtex/IEEEabrv, bibtex/bibliography}

\begin{thebibliography}{10}
\providecommand{\url}[1]{#1}
\csname url@rmstyle\endcsname
\providecommand{\newblock}{\relax}
\providecommand{\bibinfo}[2]{#2}
\providecommand\BIBentrySTDinterwordspacing{\spaceskip=0pt\relax}
\providecommand\BIBentryALTinterwordstretchfactor{4}
\providecommand\BIBentryALTinterwordspacing{\spaceskip=\fontdimen2\font plus
\BIBentryALTinterwordstretchfactor\fontdimen3\font minus
  \fontdimen4\font\relax}
\providecommand\BIBforeignlanguage[2]{{%
\expandafter\ifx\csname l@#1\endcsname\relax
\typeout{** WARNING: IEEEtran.bst: No hyphenation pattern has been}%
\typeout{** loaded for the language `#1'. Using the pattern for}%
\typeout{** the default language instead.}%
\else
\language=\csname l@#1\endcsname
\fi
#2}}

\bibitem{kemp2007challenges}
C.~C. {Kemp}, \emph{et~al.}, ``Challenges for robot manipulation in human
  environments,'' \emph{IEEE Robotics Automation Magazine}, 2007.

\bibitem{Li2020ancsh}
X.~Li, \emph{et~al.}, ``Category-level articulated object pose estimation,'' in
  \emph{CVPR}, 2020.

\bibitem{Welschehold2017humandemo}
T.~{Welschehold}, \emph{et~al.}, ``Learning mobile manipulation actions from
  human demonstrations,'' in \emph{IROS}, 2017.

\bibitem{wu2021vat}
R.~Wu, \emph{et~al.}, ``Vat-mart: Learning visual action trajectory proposals
  for manipulating 3d articulated objects,'' in \emph{ICLR}, 2022.

\bibitem{mo2021where2act}
K.~Mo, \emph{et~al.}, ``Where2act: From pixels to actions for articulated 3d
  objects,'' in \emph{ICCV}, 2021.

\bibitem{berenson2011tsr}
D.~Berenson, \emph{et~al.}, ``Task space regions: A framework for
  pose-constrained manipulation planning,'' \emph{IJRR}, 2011.

\bibitem{Burget2013wbmparticulated}
F.~Burget, \emph{et~al.}, ``{Whole-body motion planning for manipulation of
  articulated objects},'' in \emph{ICRA}, 2013.

\bibitem{Pankert2020perceptivempc}
J.~Pankert and M.~Hutter, ``Perceptive model predictive control for continuous
  mobile manipulation,'' \emph{IEEE RA-L}, 2020.

\bibitem{sleiman2021unified}
J.-P. Sleiman, \emph{et~al.}, ``A unified mpc framework for whole-body dynamic
  locomotion and manipulation,'' \emph{IEEE RA-L}, 2021.

\bibitem{gaertner2021collision}
M.~Gaertner, \emph{et~al.}, ``Collision-free mpc for legged robots in static
  and dynamic scenes,'' in \emph{ICRA}, 2021.

\bibitem{Meeussen2010dooropening}
W.~{Meeussen}, \emph{et~al.}, ``Autonomous door opening and plugging in with a
  personal robot,'' in \emph{ICRA}, 2010.

\bibitem{arduengo2019robust}
M.~Arduengo, \emph{et~al.}, ``Robust and adaptive door operation with a mobile
  robot,'' \emph{Intelligent Service Robotics}, 2021.

\bibitem{Vahrenkamp2013irm}
N.~{Vahrenkamp}, \emph{et~al.}, ``Robot placement based on reachability
  inversion,'' in \emph{ICRA}, 2013.

\bibitem{xia2020relmogen}
F.~Xia, \emph{et~al.}, ``Relmogen: Leveraging motion generation in
  reinforcement learning for mobile manipulation,'' \emph{ICRA}, 2021.

\bibitem{li2020hrl4in}
C.~Li, \emph{et~al.}, ``Hrl4in: Hierarchical reinforcement learning for
  interactive navigation with mobile manipulators,'' in \emph{CoRL}, 2020.

\bibitem{honerkamp2021learning}
D.~Honerkamp, \emph{et~al.}, ``Learning kinematic feasibility for mobile
  manipulation through deep reinforcement learning,'' \emph{IEEE RA-L}, 2021.

\bibitem{chitta2010dooropen}
S.~{Chitta}, \emph{et~al.}, ``Planning for autonomous door opening with a
  mobile manipulator,'' in \emph{ICRA}, 2010.

\bibitem{Farshidian2017brealtimeplan}
F.~Farshidian, \emph{et~al.}, ``Real-time motion planning of legged robots: A
  model predictive control approach,'' in \emph{ICHR}, 2017.

\bibitem{chiu2022collision}
J.-R. Chiu, \emph{et~al.}, ``A collision-free mpc for whole-body dynamic
  locomotion and manipulation,'' in \emph{ICRA}, 2022.

\bibitem{abbatematteo2019learning}
B.~Abbatematteo, \emph{et~al.}, ``Learning to generalize kinematic models to
  novel objects,'' in \emph{CoRL}, 2019.

\bibitem{Jain2020ScrewNet}
A.~Jain, \emph{et~al.}, ``Screwnet: Category-independent articulation model
  estimation from depth images using screw theory,'' in \emph{ICRA}, 2021.

\bibitem{zeng2020visual}
V.~Zeng, \emph{et~al.}, ``Visual identification of articulated object parts,''
  in \emph{IROS}, 2021.

\bibitem{MartinMartin2016visualpercepobj}
R.~Martin-Martin, \emph{et~al.}, ``{An integrated approach to visual perception
  of articulated objects},'' in \emph{ICRA}, 2016.

\bibitem{hausman2015active}
K.~{Hausman}, \emph{et~al.}, ``Active articulation model estimation through
  interactive perception,'' in \emph{ICRA}, 2015.

\bibitem{wong2022error}
J.~Wong, \emph{et~al.}, ``Error-aware imitation learning from teleoperation
  data for mobile manipulation,'' in \emph{CoRL}, 2022.

\bibitem{xiong2021learning}
H.~Xiong, \emph{et~al.}, ``Learning by watching: Physical imitation of
  manipulation skills from human videos,'' in \emph{IROS}, 2021.

\bibitem{xu2022umpnet}
Z.~Xu, \emph{et~al.}, ``Umpnet: Universal manipulation policy network for
  articulated objects,'' \emph{IEEE RA-L}, 2022.

\bibitem{ruhr2012openingdoors}
T.~{Rühr}, \emph{et~al.}, ``A generalized framework for opening doors and
  drawers in kitchen environments,'' in \emph{ICRA}, 2012.

\bibitem{Farshidian2017efficoptplan}
\BIBentryALTinterwordspacing
F.~Farshidian, \emph{et~al.}, ``{An efficient optimal planning and control
  framework for quadrupedal locomotion},'' in \emph{ICRA}, 2017.
\BIBentrySTDinterwordspacing

\bibitem{feller2016relaxed}
C.~Feller and C.~Ebenbauer, ``Relaxed logarithmic barrier function based model
  predictive control of linear systems,'' \emph{IEEE T-AC}, 2016.

\bibitem{pan2012fcl}
J.~Pan, \emph{et~al.}, ``Fcl: A general purpose library for collision and
  proximity queries,'' in \emph{ICRA}, 2012.

\bibitem{Han2019fiesta}
\BIBentryALTinterwordspacing
L.~Han, \emph{et~al.}, ``{FIESTA: Fast Incremental Euclidean Distance Fields
  for Online Motion Planning of Aerial Robots},'' in \emph{IROS}, 2019.
\BIBentrySTDinterwordspacing

\bibitem{Felzenszwalb2012dtsf}
P.~Felzenszwalb and D.~Huttenlocher, ``{Distance Transforms of Sampled
  Functions},'' \emph{Theory of Computing}, 2012.

\bibitem{Fankhauser2016GridMapLibrary}
P.~Fankhauser and M.~Hutter, ``{A Universal Grid Map Library: Implementation
  and Use Case for Rough Terrain Navigation},'' in \emph{Robot Operating System
  (ROS)}.\hskip 1em plus 0.5em minus 0.4em\relax Springer, 2016.

\bibitem{nvidia2020omniverse}
NVIDIA, ``Nvidia isaac sim,'' \url{https://developer.nvidia.com/isaac-sim},
  accessed: 2022-02-28.

\bibitem{Xiang2020sapien}
F.~Xiang, \emph{et~al.}, ``Sapien: A simulated part-based interactive
  environment,'' in \emph{CVPR}, 2020.

\bibitem{wu2019detectron2}
Y.~Wu, \emph{et~al.}, ``Detectron2,''
  \url{https://github.com/facebookresearch/detectron2}, 2019.

\bibitem{wright1999numerical}
J.~Nocedal and S.~J. Wright, \emph{Numerical Optimization}, 2nd~ed.\hskip 1em
  plus 0.5em minus 0.4em\relax New York, NY, USA: Springer, 2006.

\end{thebibliography}

\end{document}